\documentclass{sig-alternate}
\setcopyright{acmcopyright}
\conferenceinfo{KDD}{'16, August 13-17, 2016, San Francisco, United States.}
\usepackage{amsmath}
\usepackage{amssymb}
\usepackage{amsfonts}
\usepackage{graphicx}
\usepackage{dsfont}
\usepackage{yfonts}
\usepackage{url}
\usepackage{algorithm}
\usepackage{algorithmic}
\usepackage{subfigure}
\usepackage{comment}
\usepackage{booktabs}
\usepackage{multirow}
\usepackage{threeparttable}

\begin{document}

\title{Recruitment Market Trend Analysis with Sequential Latent Variable Models}
\numberofauthors{1}
\author{
\alignauthor Chen Zhu$^{1}$,
\hspace{1mm} Hengshu Zhu$^{1,2}$,
\hspace{1mm} Hui Xiong$^{3}$\thanks{Corresponding Author.},
\hspace{1mm} Pengliang Ding$^1$,
\hspace{1mm} Fang Xie$^1$\\
\vspace{2mm}
\affaddr{$^1$Baidu Inc.,}
\affaddr{$^2$Baidu Research-Big Data Lab,}
\affaddr{$^3$Rutgers University,}\\
\vspace{1mm}
\affaddr{\{zhuchen02, zhuhengshu, dingpengliang, xiefang\}@baidu.com,}
\affaddr{hxiong@rutgers.edu}
}

\CopyrightYear{2016}
\setcopyright{acmcopyright}
\conferenceinfo{KDD '16,}{August 13-17, 2016, San Francisco, CA, USA}
\isbn{978-1-4503-4232-2/16/08}\acmPrice{\$15.00}
\doi{http://dx.doi.org/10.1145/2939672.2939689}

\maketitle
%
%****************************************************************************************************
\begin{abstract}
Recruitment market analysis provides valuable understanding of industry-specific economic growth and plays an important role for both employers and job seekers. With the rapid development of online recruitment services, massive recruitment data have been accumulated and enable a new paradigm for recruitment market analysis. However, traditional methods for recruitment market analysis largely rely on the knowledge of domain experts and classic statistical models, which are usually too general to model large-scale dynamic recruitment data, and have difficulties to capture the fine-grained market trends. To this end, in this paper, we propose a new research paradigm for recruitment market analysis by leveraging unsupervised learning techniques for automatically discovering recruitment market trends based on large-scale recruitment data. Specifically, we develop a novel sequential latent variable model, named MTLVM, which is designed for capturing the sequential dependencies of corporate recruitment states and is able to automatically learn  the latent recruitment topics within a Bayesian generative framework. In particular, to capture the variability of recruitment topics over time, we design hierarchical dirichlet processes for MTLVM. These processes allow to dynamically generate the evolving recruitment topics. Finally, we implement a prototype system to empirically evaluate our approach based on real-world recruitment data in China. Indeed, by visualizing the results from MTLVM, we can successfully reveal many interesting findings, such as \emph{the popularity of LBS related jobs reached the peak in the 2nd half of 2014, and decreased in 2015.}
%****************************************************************************************************
\end{abstract}
\vspace{-2mm}

%****************************************************************************************************

\section{Introduction}
The scarcity of skilled talents has stimulated the global recruitment industry in the past few years.  An article from Forbes reported that, US corporations spend nearly \$ 72 billion each year on a variety of recruiting services, and the worldwide number is likely three times bigger~\cite{forbes}. Along this line, a growing challenge is to provide an effective trend analysis of recruitment market, such as forecasting recruitment demand and predicting market status. Both employers and job seekers can benefit from the study of recruitment market trends. Moreover, at the macro level, this analysis can also provide valuable understanding of industry-specific economic growth for business analysts.

With the rapid development of online recruitment services, such as Linkedin~\cite{linkedin}, Dice~\cite{dice}, and Lagou~\cite{lagou},  massive recruitment posting data have been accumulated.  For example, as of the end of 2015,
there are more than 1.3 million job positions from 100K+ companies across more than 20 Internet-related industries, such as  mobile Internet, online-to-offline (O2O), and cloud computing, available at Lagou~\cite{lagou}-a Chinese tech hiring service website. These huge data enable a new paradigm for studying recruitment market trends in a holistic and fine-grained manner.

Recruitment market analysis is a classic topic in human capital economics, where recruitment market is either treated as a factor of macro economic phenomenons or the analysis is focused on advising people to make the best job decisions in a general economic framework~\cite{romer1996advanced,shapiro1984equilibrium,varian2010intermediate}. However, previous  studies rely largely on the knowledge of domain experts and classic statistical models, and thus usually too general to capture the high variability of recruitment market (e.g., the evolution of recruitment topics). Also, these studies have limited efforts on understanding the fine-grained market trends, such as forecasting the recruitment situation for a specific company in the next few months. Therefore, it is very appealing to design a new research paradigm for recruitment market analysis through large-scale analysis on massive recruitment data. Along this line, there are some major challenges. First, how to model the intrinsically sequential dependency of recruitment states (e.g., Hiring Freeze) forming the market trend? Second, how to model the semantic relationship between the market trend and job postings from different companies? Finally, how to model the variability of recruitment postings for a long time period?

To tackle these challenges, we propose an unsupervised learning approach for recruitment market trend analysis, which can automatically discern the underlying trend of recruitment market. First, we develop a novel sequential market trend latent variable model (MTLVM), which is designed for capturing the temporal dependencies of recruitment states of companies and is able to automatically learn the latent recruitment topics and demands from  recruitment data within a Bayesian generative framework. To be more specific, we assume that the current recruitment state of a specific company is influenced by its state in previous epoch, and it will impel the company to review appropriate recruitment demands (e.g., R\&D recruitment). Then, different recruitment demands will generate different recruitment topics (e.g., experienced algorithm engineer), which will finally generate the recruitment postings. In particular, to capture the variability of recruitment topics over time, we design a hierarchical dirichlet processes for MTLVM, which can dynamically generate recruitment topics. Finally, we implement an intelligent prototype system to empirically evaluate our approach based on a real-world recruitment data set collected from China for the time period from 2014 to 2015. Indeed, by visualizing the results from MTLVM, we can successfully observe many interesting discoveries, such as \emph{the popularity of LBS related jobs reaches the peak in the 2nd half of 2014, and decreases in 2015}. Generally, the contributions of this paper can be summarized as follows.

\begin{itemize}
\vspace{-1mm}
  \item To the best of our knowledge, this paper is the first attempt to leverage unsupervised learning approach for automatically modeling the trend of recruit market. This work provides a new research paradigm for recruitment market analysis.
\vspace{-0mm}
  \item We propose a sequential latent variable model, named MTLVM, for learning the latent recruitment states, demands, and topics simultaneously. Particularly, MTLVM can dynamically generate recruitment topics by integrating hierarchical dirichlet processes.
\vspace{-0mm}
  \item We develop a prototype system for empirically evaluate our approach. Indeed, by visualizing the results obtained from MTLVM, we can successfully observe many interesting and useful findings.
\vspace{-1mm}
\end{itemize}
%****************************************************************************************************

\begin{figure}[t]
\vspace{-1mm}
\begin{center}
\includegraphics[width=8.5cm]{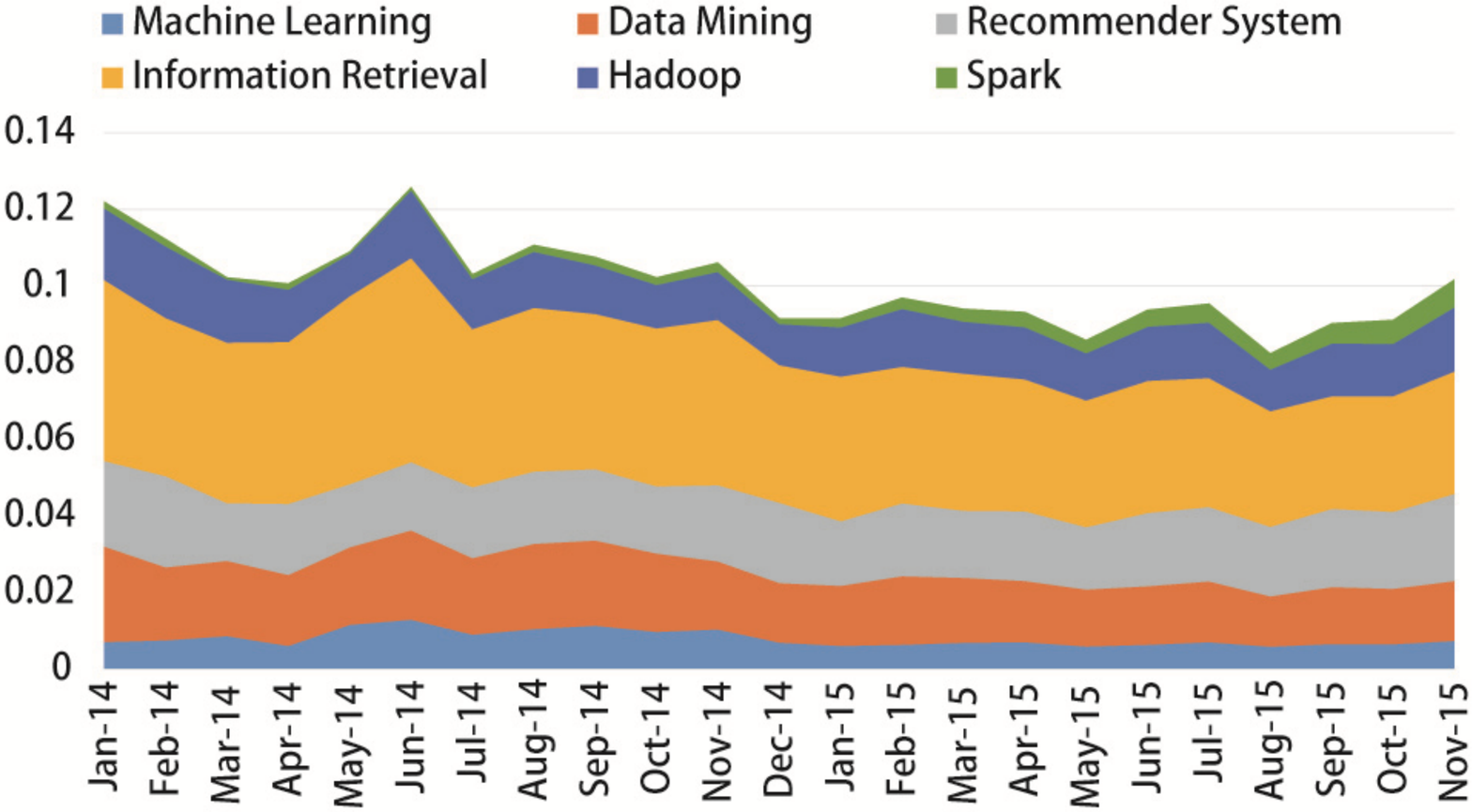}
\vspace{-5mm}
\caption{The trend of the number of job postings related to different skill requirements over two years, which indicates the demands of recruitment market.}\label{fig:over-trend}
\vspace{-5mm}
\end{center}
\end{figure}

\vspace{-1mm}
\section{Overview}
\label{sec:overview}
%****************************************************************************************************
In this section, we first introduce some preliminaries of recruitment market modeling, and then formally present the overview of our model MTLVM.

\subsection{Preliminaries}

Recent years have witnessed the rapid development of online recruitment services, which have already become the most important venue of talent seeking, especially for high-tech companies. Therefore, the job posting data from these services can help researchers better understand the trend of recruitment market from the perspectives of not only an individual company but also the whole industry.

Intuitively, different job postings can indicate different \emph{\textbf{recruitment demands}} of companies, such as R\&D related positions, which usually change over different epochs. For example, Figure~\ref{fig:over-trend} demonstrates the trend of the number of job postings related to different skill requirements during January 2014 to November 2015 based on our real-world data set.  We can observe that skill ``Information Retrieval'' becomes less popular, compared with other skills, such as ``Data Mining'', and ``Machine Learning''. Meanwhile, ``Spark'', an emerging technique for big data processing, has attracted  more and more attention during these years. Indeed, such evolution of recruitment demands is inherently determined by the change of latent \emph{\textbf{recruitment states}} of companies at different epochs, which have strong sequential dependency. For example, Alibaba, one of the largest E-Commerce companies in China, hugely enlarged the recruitment in 2014, which is followed by the recruitment state ``Hiring Freeze'' in 2015. As a result, its recruitment demands related to E-Commerce largely shrank in 2015. To capture the change of recruitment states, and model the semantic relationship between recruitment demand and state, our model MTLVM follows the Beysian latent variable model with first-order Markov assumption, where current state is only determined by the state at previous epoch.

\begin{figure}[t]
\centering
\vspace{-1mm}
\subfigure[Jan, 2014-Jun, 2014]{
\includegraphics[width=3.9cm]{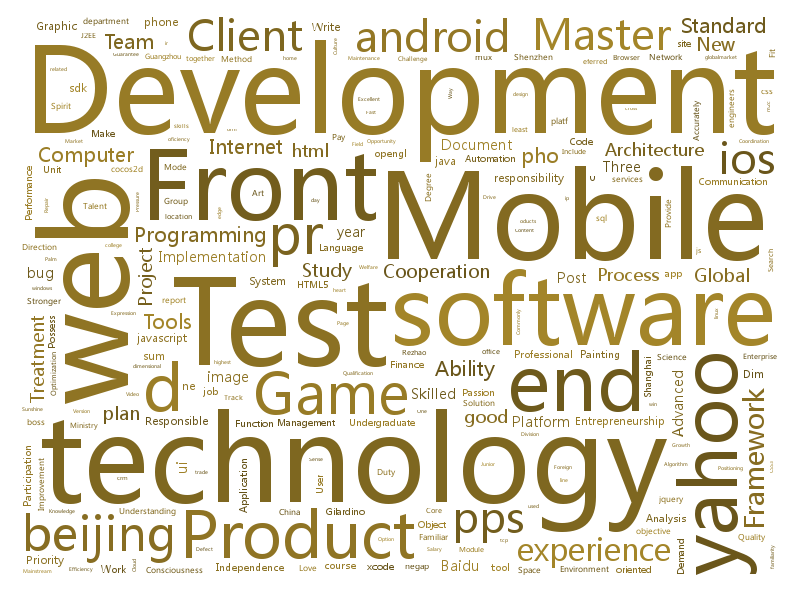}}
\subfigure[Jul, 2014-Dec, 2014]{
\includegraphics[width=3.9cm]{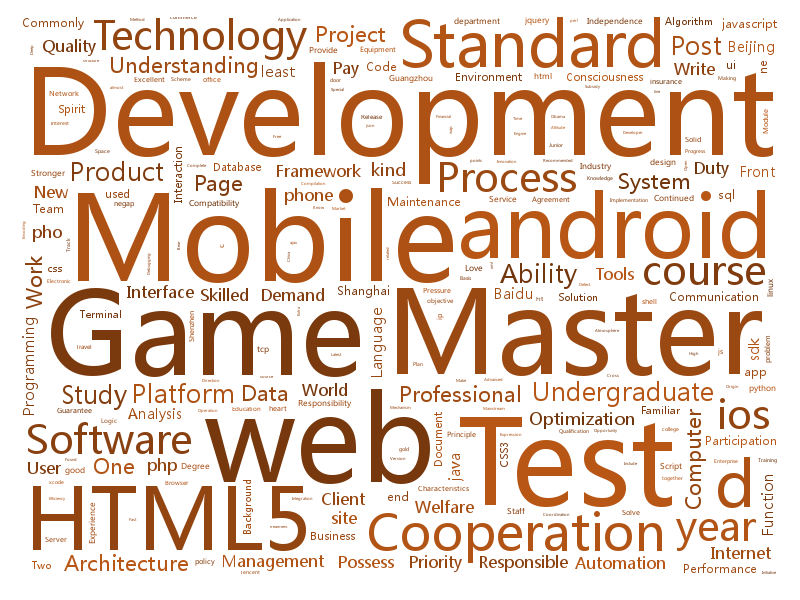}}
\subfigure[Jan, 2015-Jun, 2015]{
\includegraphics[width=3.9cm]{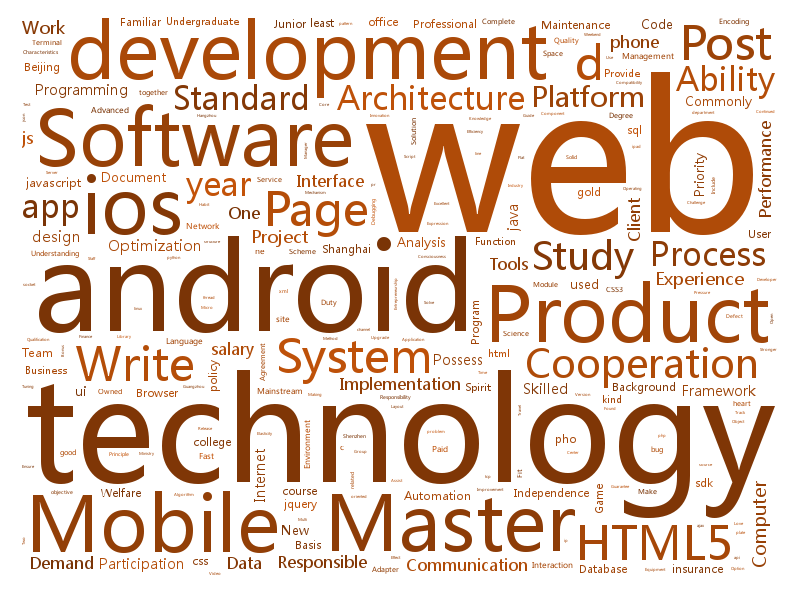}}
\subfigure[Jul, 2015-Nov, 2015]{
\includegraphics[width=3.9cm]{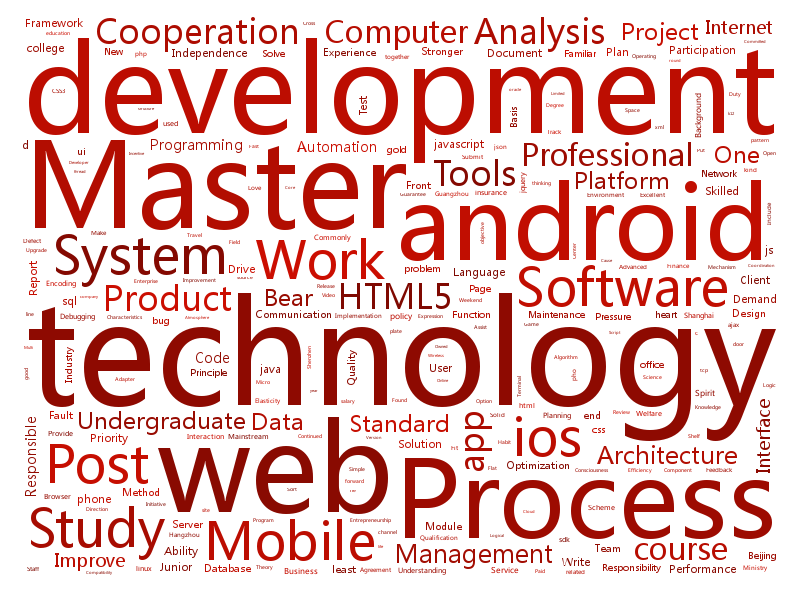}}
\vspace{-2mm}
\caption{The word cloud representation of job postings related to ``Mobile Software Engineer'' with respect to different epochs in our data set, where the size of each keyword is proportional to its frequency.}
\vspace{-4mm}
\label{data_word_cloud}
\end{figure}

Indeed, by further analyzing the descriptions of job postings, we observe that the detail of similar recruitment demands (e.g., Recruiting Mobile Software Engineer) will be influenced by the corresponding recruitment states. Thus the corresponding demands usually have high variability, and generate different \emph{\textbf{recruitment topics}}. For example, Figure~\ref{data_word_cloud} shows the word cloud representation~\footnote{All the words are originally in Chinese, and are automatically translated by a commercial translation tool~\cite{baidu-trans}.} of job postings related to ``Mobile Software Engineer'' with respect to different epochs. We can observe that, ``Mobile Game Development'' is a hot topic in the second half of 2014, while ``Android based Web Technology'' becomes popular in the first half 2015. To model the semantic relationships among recruitment states, recruitment demands, and recruitment topics, our model, MTLVM, follows the idea of Hierarchical Dirichlet Processes, an infinity version of topic modeling, to model the job postings. Therefore the topic number can be automatically determined.

\begin{figure}[t]
\begin{center}
\includegraphics[width=7cm]{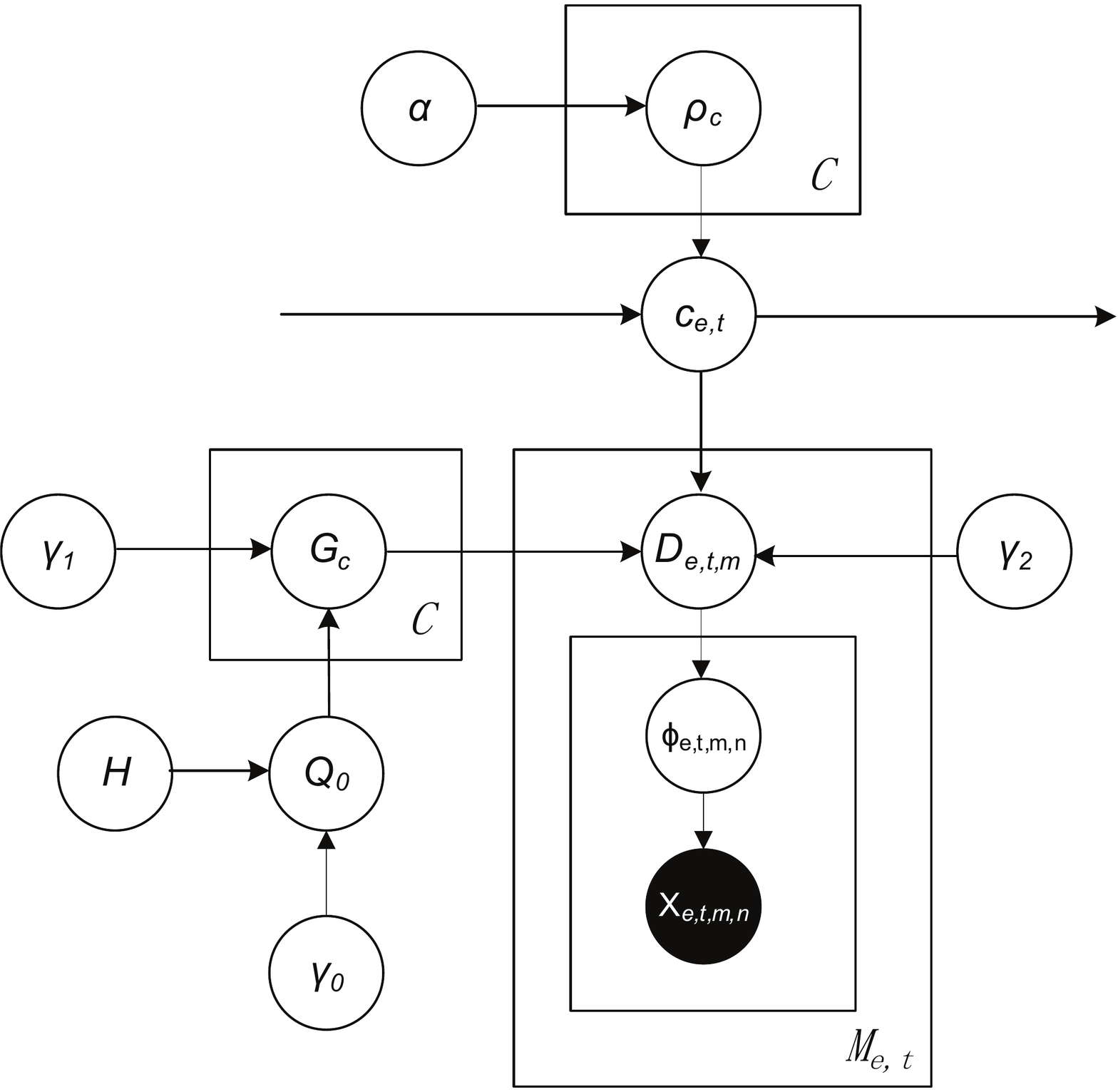}
\caption{The graphical representation of MTLVM.}\label{fig:MTLVM}
\vspace{-4mm}
\end{center}
\end{figure}

\vspace{2mm}
\subsection{Overview of MTLVM}
\vspace{1mm}
%****************************************************************************************************

Formally, we regard the $m$-th job posting of company $e$ at epoch $t$ as a bag of words $\chi_{e,t,m}=\{\chi_{e,t,m,n}\}_n$, where $\chi_{e,t,m,n}$ is the basic observation in job postings (e.g., keywords in job description), and $|\chi_{e,t,m}|=N_{e,t,m}$. For modeling the trend of recruitment market, we first divide all job postings into different data units $\{\chi_{e,t}\}_{e,t}$ with respect to companies and timestamps, which contain job postings of company $e$ at epoch $t$.

As introduced above, we assume that the current recruitment state of a specific company is influenced by its state in previous epoch, and it will impel the company to review appropriate recruitment demands. Then, different recruitment demands will generate different recruitment topics, which will finally generate the recruitment postings. Therefore, we define a parameter $c_{e,t}$ to represent the recruitment state of company $e$ at epoch $t$, where the number of unique recruitment states is $C$. Specifically, at the first epoch, the corresponding $c_{e,1}$ for all companies is sampled from a uniform distribution. For the following epochs, if company $e$ releases some job postings at previous epoch $t-1$, namely that $\chi_{e,t-1}$ exists, its current recruitment state $c_{e,t}$ is sampled from a multinomial distribution determined by previous state $c_{e,t-1}$. Otherwise, $c_{e,t}$ is drawn from the overall recruitment state at epoch $t-1$, i.e., $P(c_{e,t})=\sum_{c=1}^{C} P(c_{e,t}|c_{t-1}=c)P(c_{t-1}=c)$, where $P(c_{t-1}=c)$ is the average $P(c_{e',t-1})$ of company $e'$ who has market state $c$ at epoch $t-1$. In addition, we name the chains consisting of neighbouring $\chi_{e,t-1}$ that belong to the same company as a data chain. Therefore, if a company only occasionally releases jobs, it may have more than one data chain according to our formulation.

Furthermore, we define the generative process of job posting $\chi_{e,t,m} \in \chi_{e,t}$ of company $e$ at epoch $t$ as follows. First, a recruitment demand $D_{e,t,m}$ is generated from the latent factor $G_c$ which is sampled from a Dirichlet Process and determined by the current recruitment state $c_{e,t}$. Then, we sample a recruitment topic $\phi_{e,t,m,n}$ from the demand $D_{e,t,m}$ for each observation $\chi_{e,t,m,n}$. Finally, each observation $\chi_{e,t,m,n}$ is generated from a multinomial distribution determined by corresponding topic $\phi_{e,t,m,n}$. Specifically, Figure~\ref{fig:MTLVM} shows the graphical representation of MTLVM.

\section{Modeling the Trend of Recruitment Market}
\label{sec:model}
%****************************************************************************************************
In this section, we will introduce the technical details of our model MTLVM. And we illustrate the important mathematical notations in Table~\ref{tab:notation}.

%****************************************************************************************************
\subsection{Model Inference}
\label{sec:model1}
%****************************************************************************************************
According to the introduction in Section~\ref{sec:overview}, we can summarize the parameterizations of MTLVM as follows,
\begin{align*}
\rho_c|\alpha & \sim \text{Diri}(\alpha),\\
c_{e,t}|\{\rho_c\}_{c=1}^{C},c_{e,t-1} & \sim \text{Multi}(\rho_{c_{e,t-1}}),\\
Q_0|H,\gamma_0 & \sim \text{DP}(\gamma_0,H),\\
G_c|Q_0,\gamma_1 & \sim \text{DP}(\gamma_1,Q_0),\\
D_{e,t,m}|\{G_c\}_{c=1}^{C},\gamma_2,c_{e,t} & \sim \text{DP}(\gamma_2,G_{c_{e,t}}),\\
\phi_{e,t,m,n}|G_{e,t,m} & \sim D_{e,t,m},\\
\chi_{e,t,m,n}|\phi_{e,t,m,n} & \sim \text{Multi}(\phi_{e,t,m,n}).
\end{align*}

Following the above parameterizations, we can get the joint probability distribution of $\chi$ and $c$ as
\begin{equation}
\label{equ:chi&c}
P(\chi,c|\Lambda)=\prod_{e,t}\Big(P(c_{e,t}|c_{e,t-1},\Lambda) P(\chi_{e,t}|\Lambda,c_{e,t})\Big),
\end{equation}
where $\Lambda$ is a set of hyper-parameters, including $\alpha$, $H$, $\gamma_0$, $\gamma_1$, $\gamma_2$, and $c_{0}$. Specifically, $c_0$ is a default initial recruitment state and $\rho_{c_0}$ is fixed to $(1/C,...,1/C)$. Indeed, the above equation can be divided into two parts, i.e., $P(c_{e,t}|c_{e,t-1},\Lambda)$ that follows a multinomial distribution $\text{Multi}(\rho_{c_{e,t-1}})$, and $P(\chi_{e,t}|\Lambda,c_{e,t})$ that follows Dirichlet Processes, respectively. Specifically, $P(\chi_{e,t}|\Lambda,c_{e,t})$ can be computed by
\begin{eqnarray}
\small
\label{chi_et}
\nonumber P(\chi_{e,t}|\Lambda,c_{e,t})= \prod_{m=1}^{M_{e,t}}\Big(P(D_{e,t,m}|\Lambda,G_{c_{e,t}})\times \\
 \prod_{n=1}^{N_{e,t,m}}P(\phi_{e,t,m,n}|\Lambda,D_{e,t,m})P(\chi_{e,t,m,n}|\phi_{e,t,m,n})\Big).
\end{eqnarray}
Therefore, the objective of learning MTLVM is to find a set of optimal parameters in $P(c_{e,t}|c_{e,t-1},\Lambda)$, $P(D_{e,t,m}|\Lambda,G_{c_{e,t}})$, $P(\phi_{e,t,m,n}|\Lambda,D_{e,t,m})$, and $P(\chi_{e,t,m,n}|\phi_{e,t,m,n})$, which can maximize the probability of Equation~\ref{equ:chi&c}. In this paper, we propose a two-step framework to learn our model by a Gibbs Sampling method.

In the first step, we introduce how to optimize the transition matrix $\rho$, which is constituted of $P(c_{e,t}|c_{e,t-1},\Lambda)$, given $P(\chi_{e,t}|\Lambda,c_{e,t})$. Depending on equation~\ref{equ:chi&c}, we could get the conditional distribution of $c_{e,t}$

\begin{equation}
\begin{aligned}
P(c_{e,t}=c|c^{-e,t},\chi, \Lambda) = \frac{P(c_{e,t}=c,\chi_{e,t}|\chi^{-e,t}, \Lambda)}{P(\chi_{e,t}|c^{-e,t}, \chi^{-e,t}, \Lambda)}\\
\propto P(\chi_{e,t}|c_{e,t}=c, \Lambda)P(c_{e,t}=c|c^{-e,t}, \Lambda).\\
\end{aligned}
\end{equation}
Since $P(\chi_{e,t}|c_{e,t}=c, \Lambda)$ is given, the only challenge is to calculate $P(c_{e,t}=c|c^{-e,t}, \Lambda)$. Here we follow the inference in~\cite{goldwater2007fully} and give it directly.
\begin{eqnarray}
\nonumber & P(c_{e,t}=c|c^{-e,t},\chi, \Lambda) = P(\chi_{e,t}|c_{e,t}=c, \Lambda)\cdot \\
& \frac{ q_{(c_{e,t-1},c)}^{-e,t}+\alpha }{ q_{(c_{e,t-1})}^{-e,t}+C\alpha }\frac{ q_{(c,c_{e,t+1})}^{-e,t}+I(c_{t-1}=c=c_{t+1})+\alpha }{ q_{(c)}^{-e,t}+I(c_{t-1}=c)+C\alpha },
\end{eqnarray}
where $q_{(c)}^{-e,t}$ means the number of recruitment states $c$ appearing except $c_{e,t}$, and $q_{(c_{e,t-1},c)}^{-e,t}$ means the number of pair $c_{e,t-1},c$ appearing except $c_{e,t}$.

\begin{table}[t]
 \caption{Mathematical Notations.}\label{tab:notation}
 \small
 \centering
 \begin{tabular}{c p{6.2cm}}
  \toprule
  Symbol & Description \\
  \midrule
  $\chi_{e,t,m,n}$   &  The $n$-th tokens at $m$-th job posting of company $e$ at epoch $t$.  \\
  $\chi_{e,t,m}$   &  The $m$-th job posting of company $e$ at epoch $t$.  \\
  $\chi_{e,t}$   &  The observation unit containing job postings of company $e$ at epoch $t$.  \\
  $\chi$   &  The entire data set of job postings.  \\
  $c_{e,t}$   &  The recruitment state of company $e$ at epoch $t$.  \\
  $\alpha$   &  The hyperparameter of the Dirichlet prior on $\rho_c$.  \\
  $\rho$   &  The transition matrix of recruitment state $c$.  \\
  $G_c$   &  The probability measure representing the recruitment strategy of state $c$.  \\
  $D_{e,t,m}$   &  The probability measure representing the recruitment demand of $\chi_{e,t,m}$.  \\
  $\phi_{e,t,m,n}$   &  The recruitment topic of the $n$-th tokens at $m$-th job posting of company $e$ at epoch $t$.  \\
  $Q_0$   &  The base measure for Dirichlet Process generating $G_c$.  \\
  $\gamma_0$   &  The concentration parameter for Dirichlet Process generating $Q_0$.  \\
  $\gamma_1$   &  The concentration parameter for Dirichlet Process generating $G_c$.  \\
  $\gamma_2$   &  The concentration parameter for Dirichlet Process generating $D_{e,t,m}$.  \\
  $H$   &  The base measure for Dirichlet Process generating $Q_0$.  \\
  $\Lambda$    &  A set of hyperparameters, including $\alpha$, $H$, $\gamma_0$, $\gamma_1$, $\gamma_2$, and $c_{0}$.  \\
  $c_{0}$      & The default initial recruitment state.  \\
  $N_{e,t,m}$  &  The number of tokens at $m$-th job posting of company $e$ at epoch $t$.  \\
  $M_{e,t}$    &  The number of job postings of company $e$ at epoch $t$.  \\
  $C$   &  The number of unique recruitment states.  \\
  \bottomrule
 \end{tabular}
 \vspace{-4mm}
\end{table}

In the second step, we introduce how to compute the parameters related to Dirichlet Process in Equation~\ref{chi_et}. Indeed, this task can be regarded as an analog of the Chinese Restaurant Process (CRP)~\cite{aldous1985exchangeability}, and the metaphor can be explained as follows. We have $C$ cuisine styles (i.e., recruitment state) and a franchise (i.e., company) with $M_{e,t}$ restaurants (i.e., job postings). Everyday, the franchise will change its cuisine style according to the cuisine styles on last day. In Particular, the menus of different restaurants may be different, even if they share the same cuisine style. At each table of each restaurant, the dish (i.e., topic) is determined by the first customer (i.e., the basic observation in job postings) sitting there, and it is shared among all customers who sit at that table. When a new customer enters the restaurant, she can sit at an occupied table or a new table. If she chooses a new table, she can order a new dish from the menu. According to the above metaphor of CRP, we can easily obtained a Gibbs sampling scheme for posterior sampling given $\chi$~\cite{teh2006hierarchical}. The detailed definition and inference can be found in Appendix.

%****************************************************************************************************
\subsection{The Application of MTLVM}
%****************************************************************************************************
After learning stage, MTLVM can be used for predicting the future trend of recruitment market, e.g., recruitment states, demands, topics, and basic observations. Specifically, given a company $e$, we can estimate its current recruitment state $c_{e,t}$ by
\begin{eqnarray}
c_{e,t}=\arg\max_cP(c|\chi_{e,t}, c_{e,t-1},\rho,\Lambda),
\end{eqnarray}
where $\rho$ is the transition matrix, and
\begin{eqnarray}
\nonumber P(c_{e,t}|\chi_{e,t}, c_{e,t-1},\rho,\Lambda) \propto P(c_{e,t},\chi_{e,t}|c_{e,t-1},\rho,\Lambda)\\
                                = P(\chi_{e,t}|c_{e,t},\Lambda)P(c_{e,t}|c_{e,t-1},\rho).
\end{eqnarray}
Therefore, the probability of recruitment state at the next epoch $c_{e,t+1}$ can be obtained by $P(c_{e,t+1})=\text{Multi}(\rho_{c_{e,t}})$. Furthermore, the recruitment topics can be obtained in the same way introduced in Section~\ref{sec:model1}. Thus, the probability of a basic observation $\chi_{e,t+1,k}$ (e.g., keywords in job description) from company $e$ appearing at epoch $t+1$ can be computed by
\begin{equation}\label{equ:pred}
P(\chi_{e,t+1,k}|\Lambda)=\sum_{c_{e,t+1}=1}^{C}P(\chi_{e,t+1,k}|c_{e,t+1},\Lambda)P(c_{e,t+1}),
\end{equation}
where $P(\chi_{e,t+1,k}|c_{e,t+1})$ can be obtained by Equation~\ref{chi_et}.

\section{Experiments}
\label{sec:exp}
%****************************************************************************************************
\begin{figure}[t]
\begin{center}
\includegraphics[width=8cm,height=6cm]{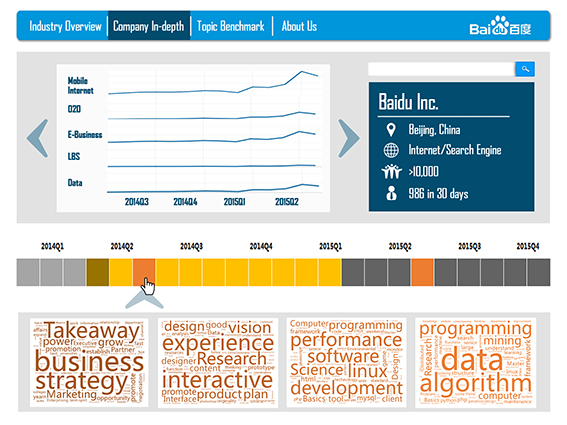}
\caption{A screenshot of our system for recruitment market analysis.}\label{fig:demo}
\vspace{-4mm}
\end{center}
\end{figure}
In this section, we will study the performance of our model MTLVM on a huge data set collected from a major online recruitment website in China.

Furthermore, we have developed a web-based prototype system to empirically evaluate our model. This system can visualize the results of our model, provide in-depth analysis of recruitment market analysis, and help people understand the high variability of recruitment market. Figure~\ref{fig:demo} shows a screenshot of this prototype system. In this system, we show the trend analysis of the entire recruitment market and the detail evolution of recruitment state of companies. All of following visualization results in this section can be obtained by this prototype system.

\begin{figure*}[!t]
\centering
\subfigure[] {\label{fig:unit_num_by_epoch}
\includegraphics[width=4.5cm]{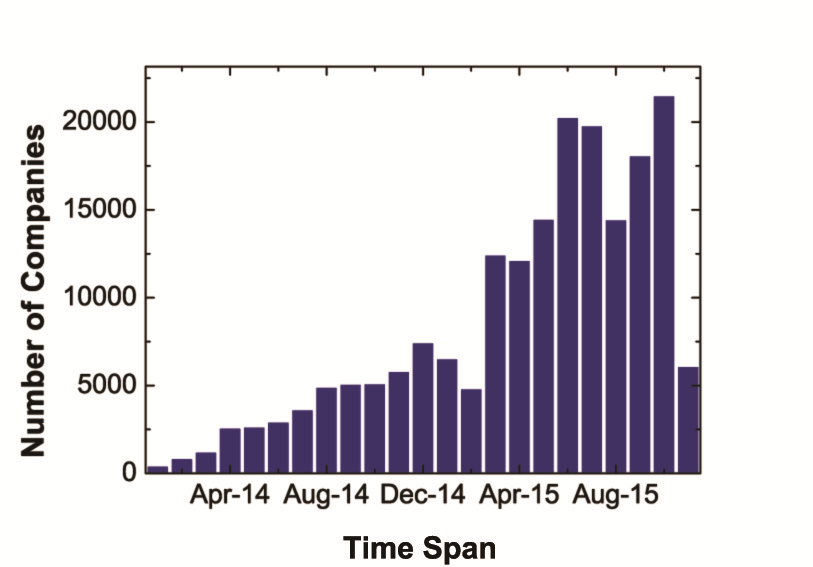}
}
\hspace{-8mm}
\subfigure[] {\label{fig:doc_num_by_company}
\includegraphics[width=4.5cm]{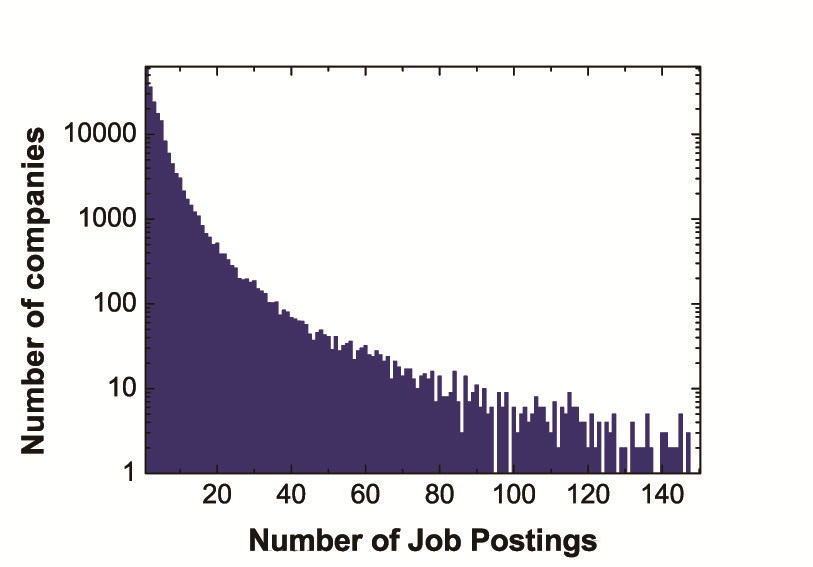}
}
\hspace{-8mm}
\subfigure[] {\label{fig:doc_num_by_unit}
\includegraphics[width=4.5cm]{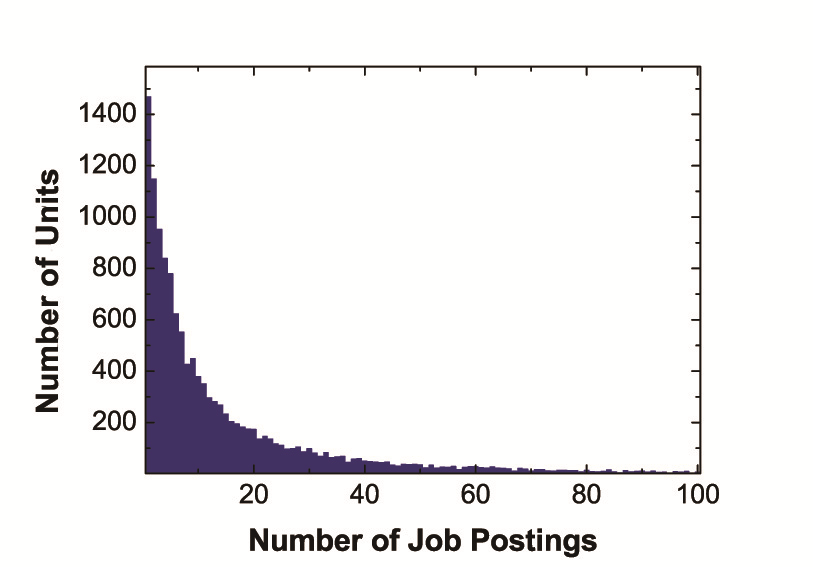}
}
\hspace{-8mm}
\subfigure[] {\label{fig:units_num_by_chain}
\includegraphics[width=4.5cm]{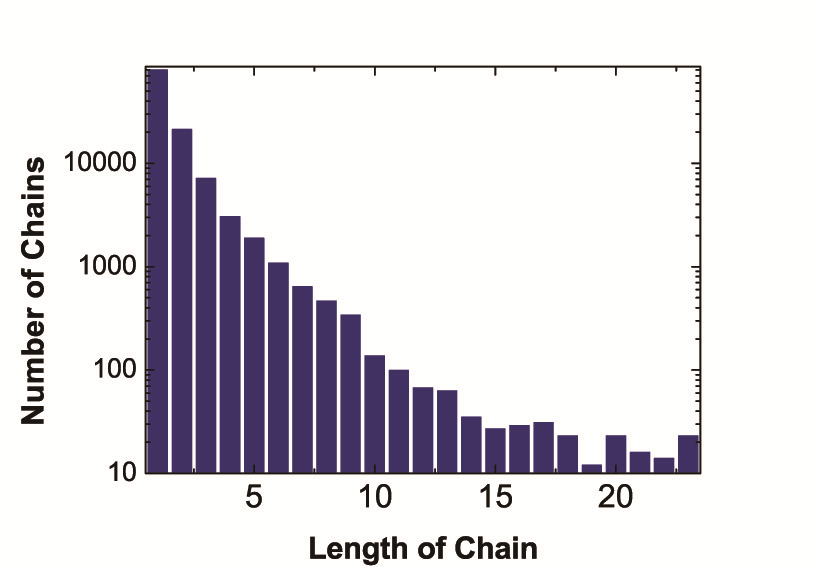}
}
\vspace{-2mm}
\caption{The distribution of (a) the number of companies that release job posting in different epochs, (b) the number of companies with respect to the number of their job postings, (c) the number of data units with respect to the number of contained job postings, and (d) the number of data chains with respect to their length.}
\vspace{-4mm}
\label{data_decs}
\end{figure*}

\begin{table}[th]
 \caption{The statistics of our real-world data set.}
 \centering
 \begin{tabular}{l r r}
  \toprule
   & Raw Data & Filtered Data \\
  \midrule
\# job postings & 934,761 & 257,166 \\
\# unique companies & 68,302 & 997 \\
\# data units & 191,549 & 13,209 \\
\# data chains & 116,392 & 2,557 \\
  \bottomrule
 \end{tabular}\label{tab:data_stat}
 \vspace{-2mm}
 \
\end{table}
%****************************************************************************************************
\vspace{2mm}
\subsection{Data Set and Experimental Setup}
\vspace{1mm}
The data set used in our experiments is collected from a major online recruitment website in China and contains 934,761 job postings from 68,302 companies released from January 2014 to November 2015. Specially, Figure~\ref{fig:unit_num_by_epoch} to \ref{fig:units_num_by_chain} demonstrate some statistics of our data set. As mentioned above, ``data unit'' in Figure~\ref{fig:doc_num_by_unit} means a job posting set $\chi_{e,t}$ released by company $e$ at epoch $t$ and ``data chain'' means chains consisting of neighbouring $\chi_{e,t-1}$ that belong to the same company. From these statistics we can observe that most of companies only randomly release very few job postings, and therefore cannot represent the trend of recruitment market. To avoid such bias, we only conserve companies which have released more than 100 job postings. Table~\ref{tab:data_stat} shows the detailed statistics of our raw data set and the filtered data set. By the above pre-process, we filtered about 72\% original data; and the number of companies declined by 98.6\%. However, the average number of job postings per company increases from 13.7 to 258, and the average length of chain also increases from 1.65 to 5.16, which make it more reasonable for training MTLVM.

In particular, in each job posting, the keywords in job description (e.g., job responsibility, skill requirements) are treated as basic observations, and all stop words are removed to guarantee the modeling performance. Note that, our model is trained with original Chinese words, and all experimental results were translated into English by a commercial translation tool~\cite{baidu-trans} for facilitating demonstration.

In the following subsections, we will comprehensively study the performance of MTLVM in term of trend analysis (e.g., learning recruitment states and recruitment topics). Specially, following \cite{teh2006hierarchical}, we set a symmetric Dirichlet distribution with parameters of 0.5 for the prior $H$ over topic distributions. For simplicity, $\gamma_0$, $\gamma_1$, and $\gamma_2$ are set to 1 directly, and another hyperparameter $\alpha$ is also be set to 1 empirically.

%****************************************************************************************************

\vspace{2mm}
\subsection{Evaluation of Recruitment Topics}
\vspace{1mm}
How to quantitatively evaluate the performance of latent variable models is always an open problem. Although perplexity and held-out likelihood are common measures for evaluating prediction results, they cannot demonstrate how coherent and meaningful the latent factors (e.g., recruitment states and topics) are.

Therefore, in this paper, we follow the measures introduced in~\cite{zhang2015dynamic}, which is inspired by~\cite{xie2013integrating,chang2009reading}, for evaluating MTLVM. Specifically, we picked up top 10 keywords for each learned recruitment topic and asked 4 senior experts of human resource to evaluate its value. These experts are first required to judge whether this topic is \emph{valuable}. If a topic is \emph{valuable}, they need continue to judge how many words are \emph{relevant} in the top 10 keyword list. Based on these manually labeled results, the metrics Validity Measure~(VM) and Coherence Measure~(CM) are defined as
\begin{equation*}
VM=\frac{\text{\# of \emph{valid} topics}}{\text{\# of topics}}, CM=\frac{\text{\# of \emph{relevant} words}}{\text{\# of words in \emph{valid} topics}}.
\end{equation*}

Besides, to evaluate how the number of recruitment states affects the performance, we train MTLVM with different settings of state number, i.e., $C=5,10,20$ respectively. Furthermore, we select widely used topic model Latent Dirichlet Allocation (LDA)~\cite{blei2003latent} as baseline. After convergence, the numbers of topic in all our models, i.e., $K$, are automatically determined as about 100. Therefore, we set $K$ as 100 for LDA.

Table~\ref{tab:VMCM} shows the average results of VM and CM.  We can observe that in terms of VM, both of MTLVM~(C=20) and MTLVM~(C=10) outperform LDA a lot, and MTLVM~(C=20) has the best performance. In terms of CM, the performance of MTLVM~(C=10) is the best, while that of MTLVM~(C=20) is worse than LDA. It may be because that too many states will make the model relatively sparse, and thus will make relevant words scattered in different topics. In particular, the performance of MTLVM~(C=5) is the worst, which may be because that few states cannot accurately describe the market trend well. Overall, since MTLVM~(C=10) has the most balanced results, we set state number $C=10$ in all of following experiments.

\begin{table}[t]
\vspace{-0mm}
 \caption{Average VM/CM comparison.}
 \centering
 \begin{tabular}{l r r r}
  \toprule
   & K & VM & CM \\
  \midrule
    MTLVM~(C=5)  & 105  & 0.464     & 5.044 \\
    MTLVM~(C=10) & 100   & 0.682     & \textbf{6.783} \\
    MTLVM~(C=20) & 118   & \textbf{0.688}     & 6.279 \\
    LDA          & 100   & 0.637    & 6.722 \\
  \bottomrule
 \end{tabular}\label{tab:VMCM}
 \vspace{-2mm}
\end{table}

\begin{figure}[!t]
\begin{center}
\vspace{-0mm}
\includegraphics[width=8cm]{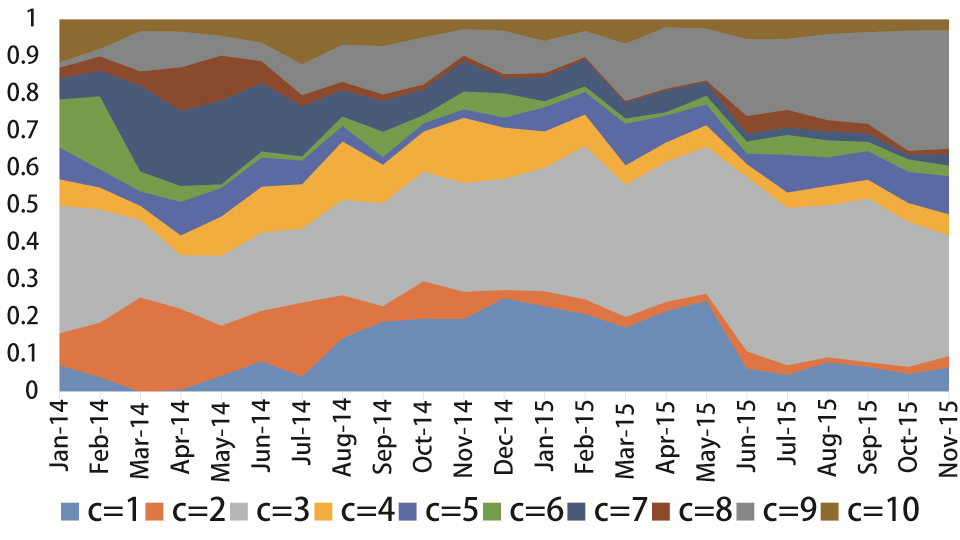}
\vspace{-2mm}
\caption{The trend of popularity of recruitment states discovered by our model over time.}\label{fig:state&time}
\vspace{-5mm}
\end{center}
\end{figure}

\subsection{Evaluation of Recruitment States}
%****************************************************************************************************
Here, we will empirically evaluate the learned recruitment states from several aspects.

\begin{figure}[t]
\begin{center}
\includegraphics[width=8cm]{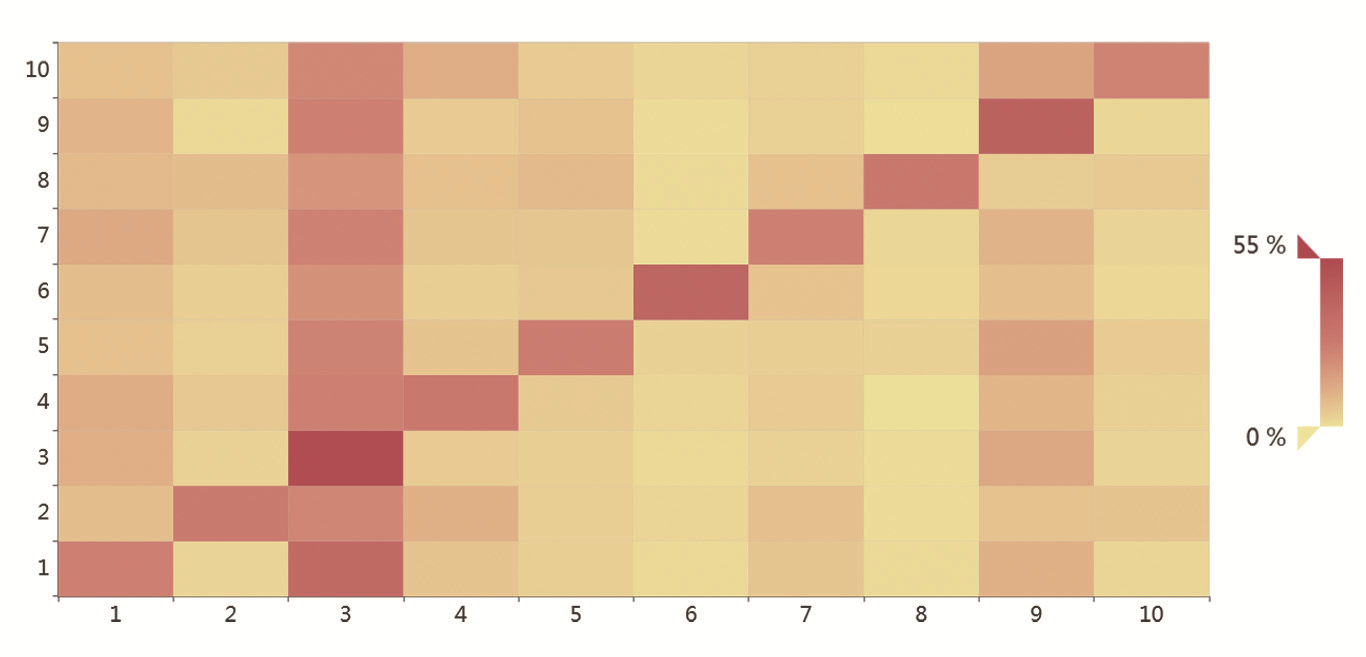}
\vspace{-4mm}
\caption{The transition matrix of recruitment states, where the $i,j$ element means the transition probability from $i$-th state to $j$-th state, and deeper color means higher probability.}\label{fig:heatmap}
\vspace{-6mm}
\end{center}
\end{figure}

Figure~\ref{fig:state&time} shows the trend of popularity of recruitment states discovered by MTLVM over time. It is obvious that these states change over time dramatically. Specifically, state \#3 kepng a relative high popularity over the long period, while popularity of \#6 is always low. Meanwhile, the popularity of state \#9 is rising dramatically since February 2015. Several other states, such as state \#2, \#7, and \#10, represent totally opposite trends. Furthermore, Figure~\ref{fig:heatmap} shows the transition matrix of recruitment states, where the element of $i$-th row and $j$-th column represents the transition probability from state $i$ to state $j$. We can observe that, all states have the highest transition probabilities to themselves, which is due to the momentum of recruitment market. Also, the color of $3$-th, and $9$-th columns is relatively deeper, which indicates the importance of states \#3, and \#9. All of above results show that our model MTLVM has the ability to capture the high variability of recruitment market by discovering these latent recruitment states.

\subsubsection{Recruitment State Inspection}

\begin{figure*}[t!]
\centering
\subfigure[state \#3] {
\begin{minipage}[b]{0.23\textwidth}
\includegraphics[width=1\textwidth]{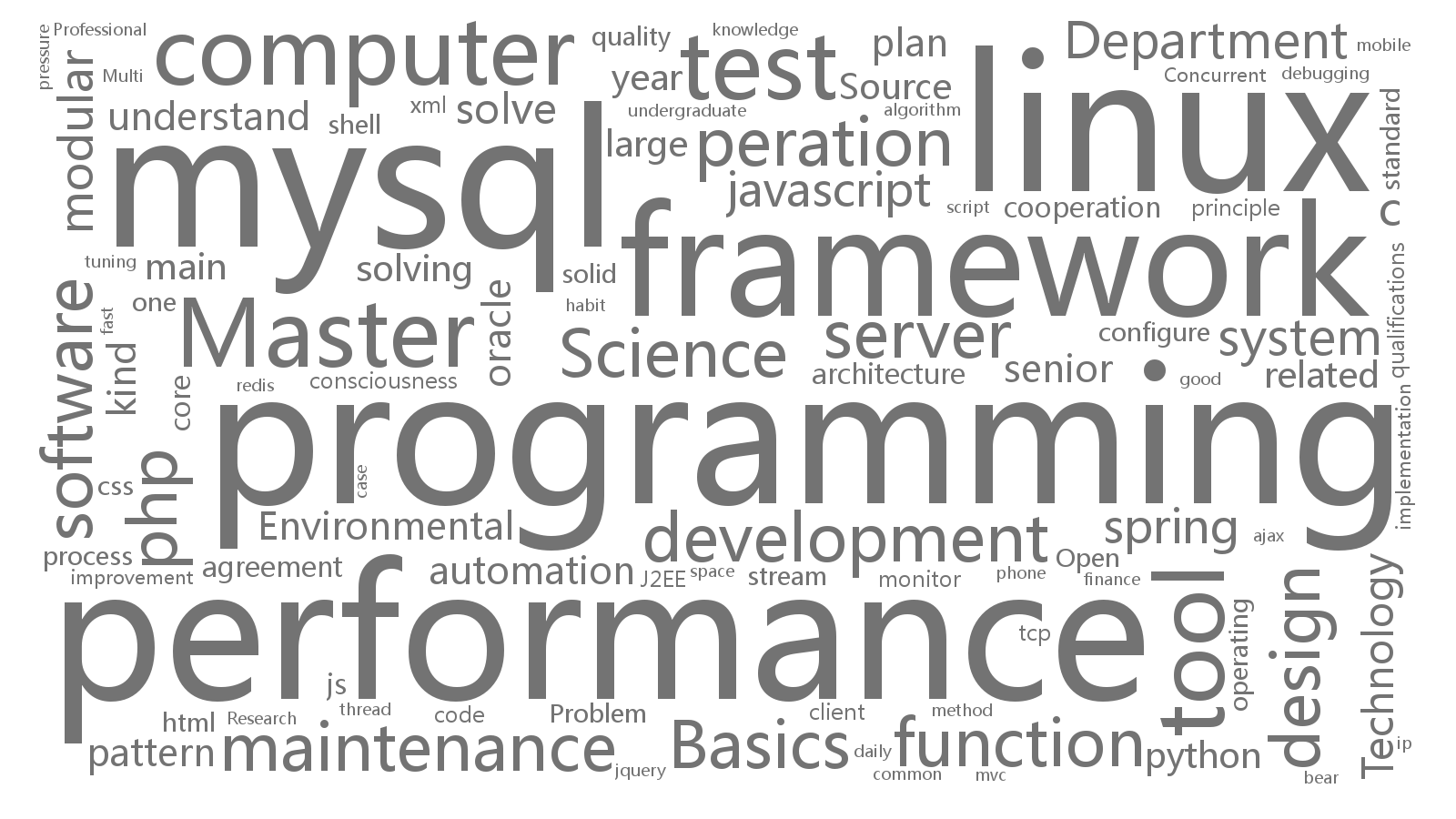} \\
\includegraphics[width=1\textwidth]{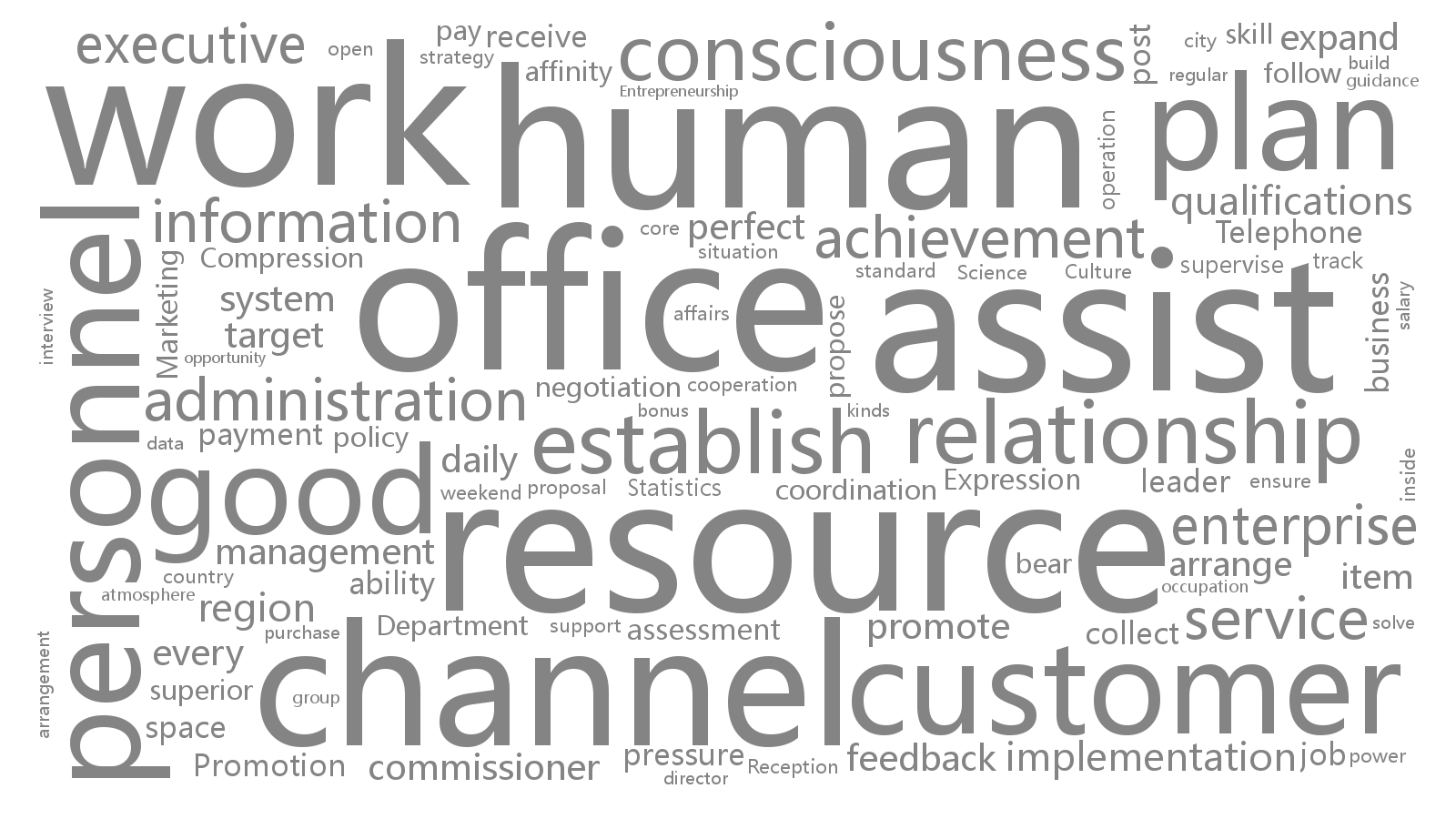} \\
\includegraphics[width=1\textwidth]{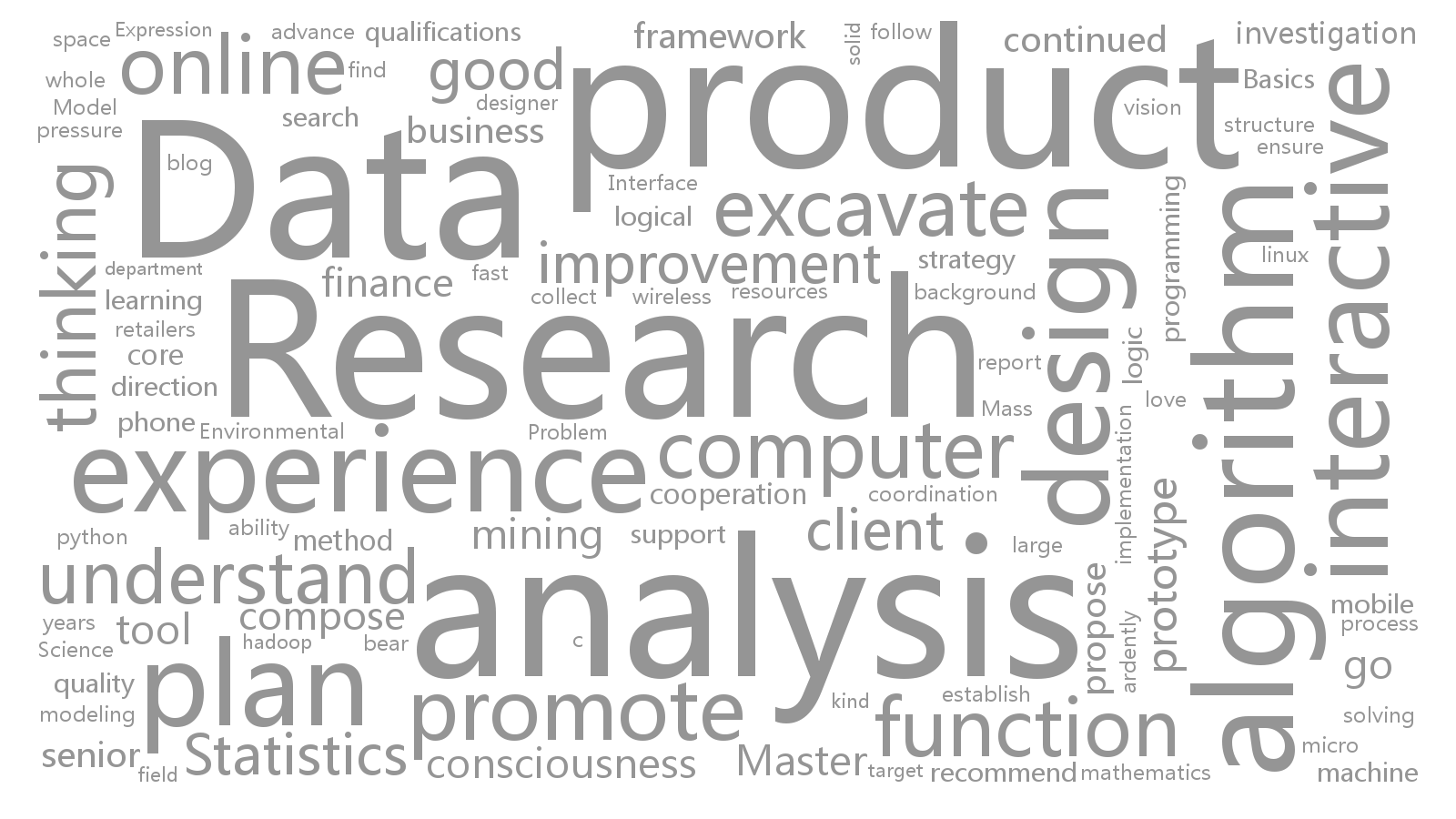} \\
\includegraphics[width=1\textwidth]{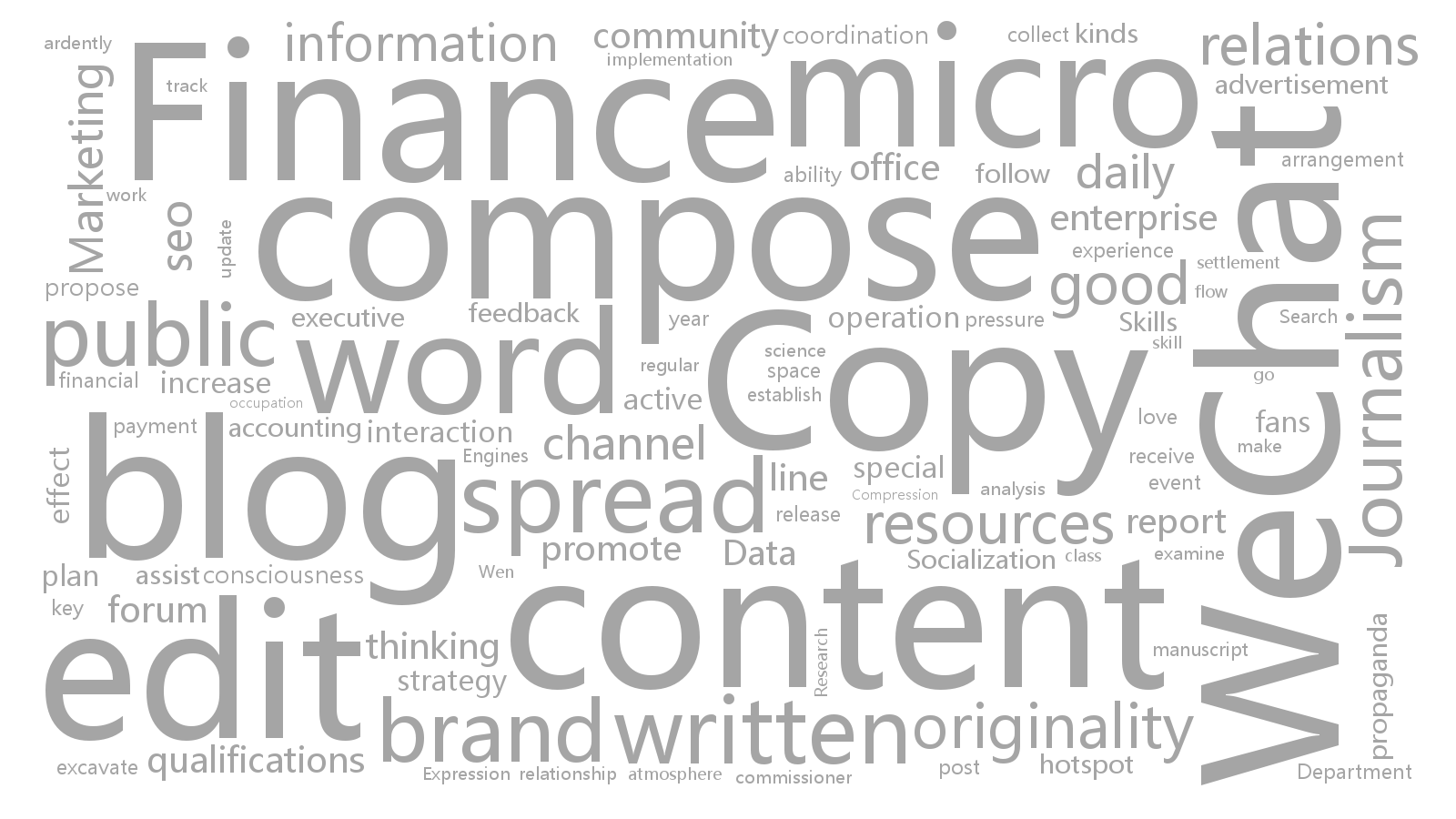}
\end{minipage}
}
\subfigure[state \#4] {
\begin{minipage}[b]{0.23\textwidth}
\includegraphics[width=1\textwidth]{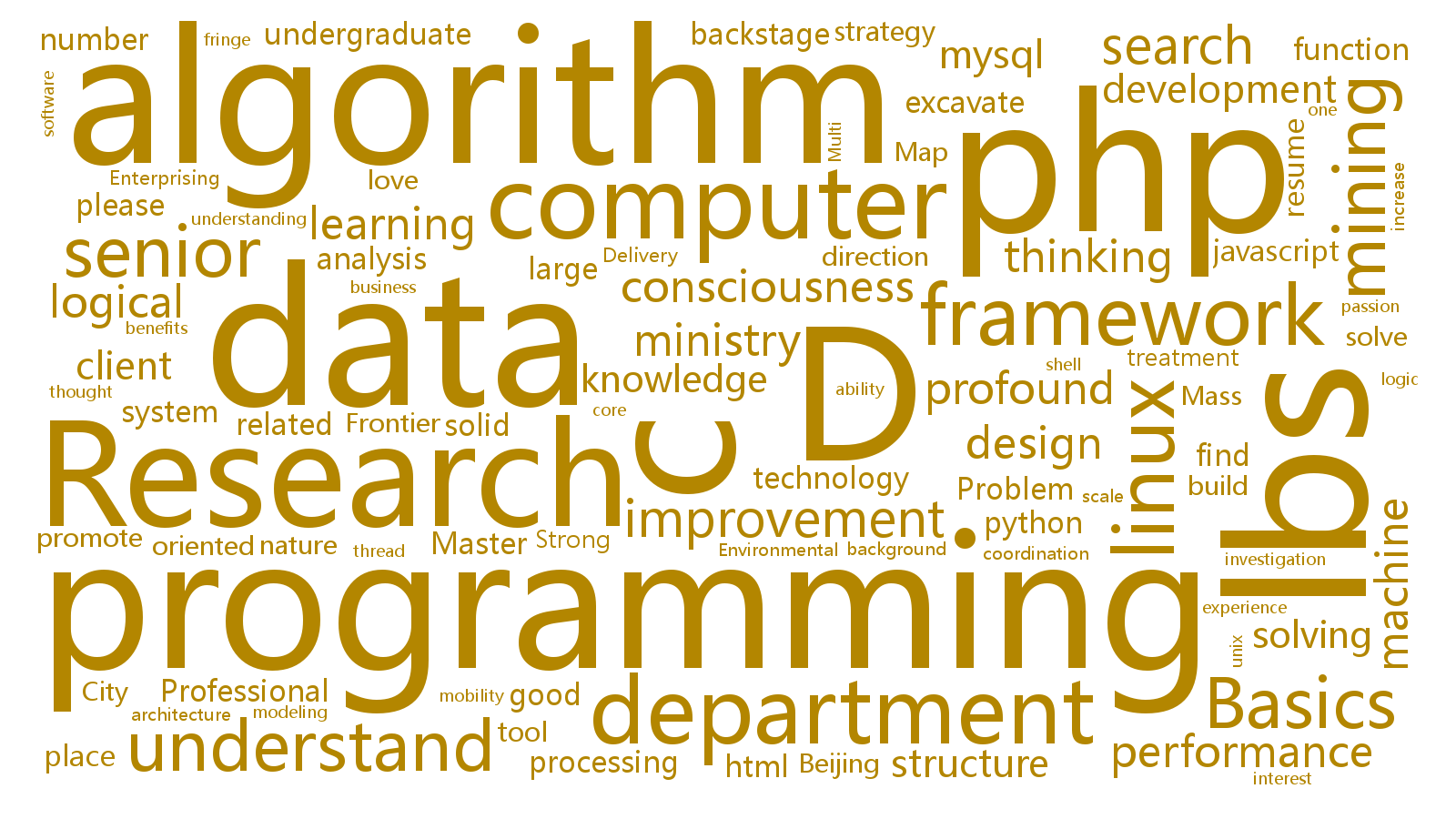} \\
\includegraphics[width=1\textwidth]{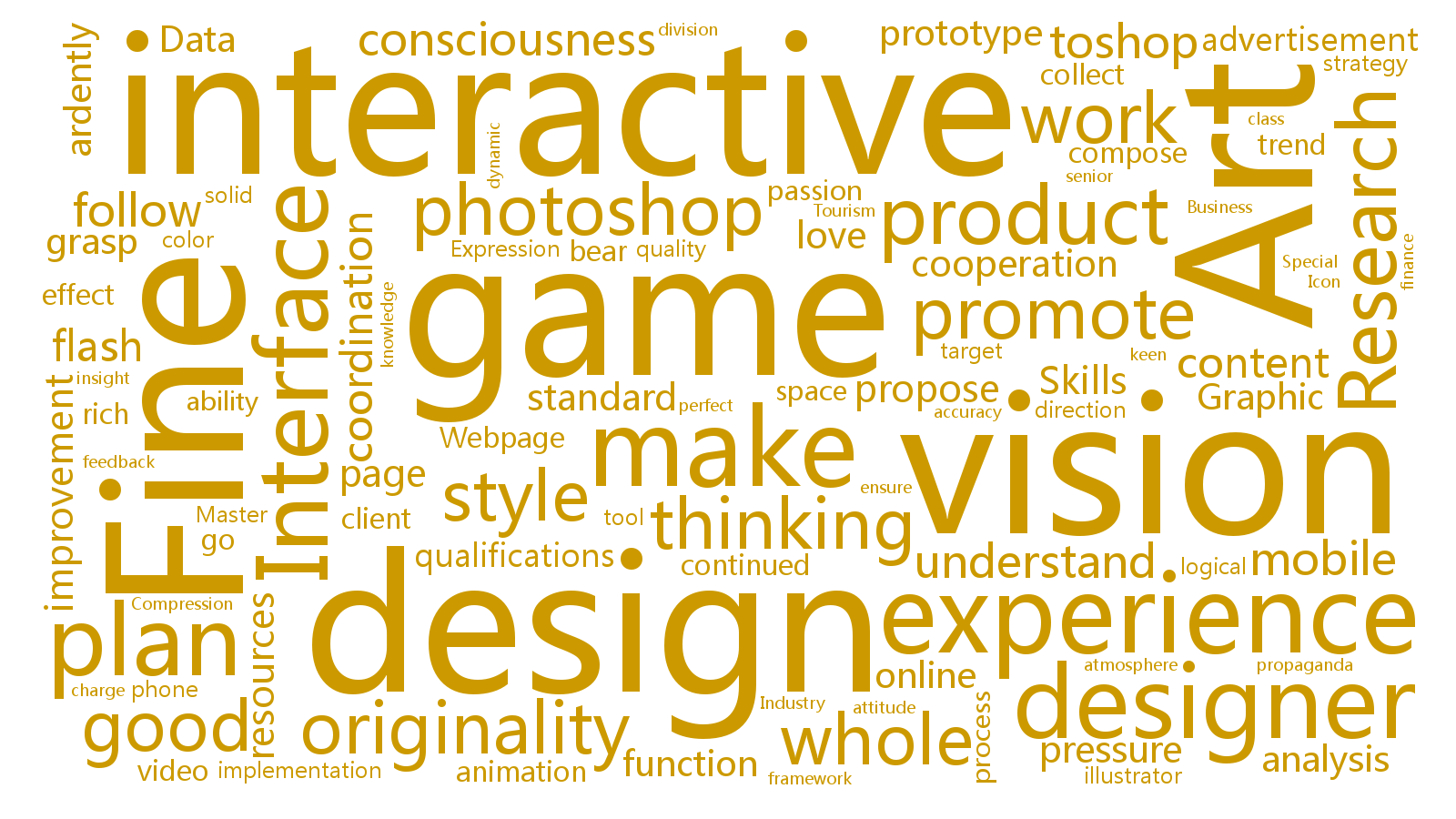} \\
\includegraphics[width=1\textwidth]{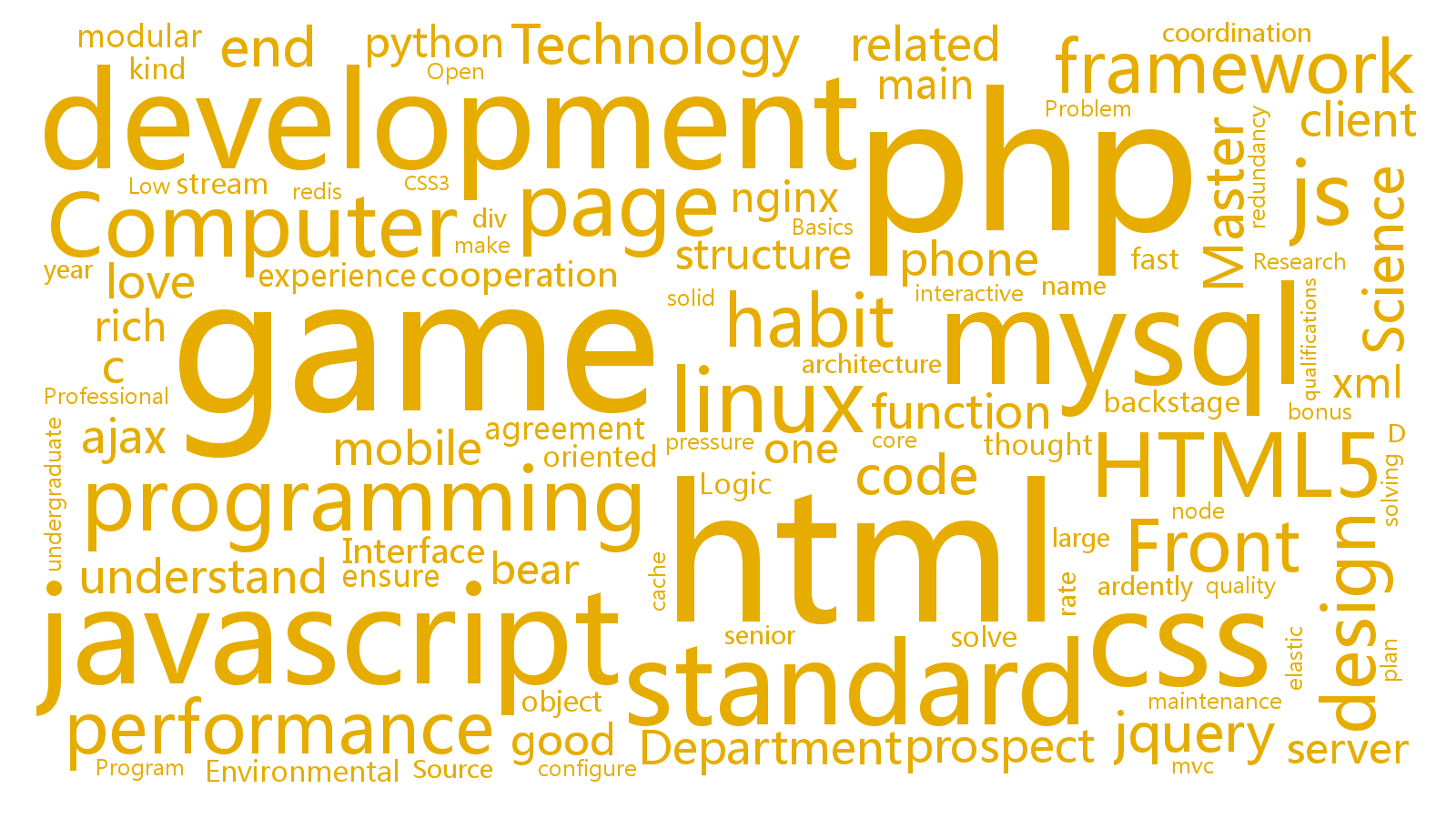} \\
\includegraphics[width=1\textwidth]{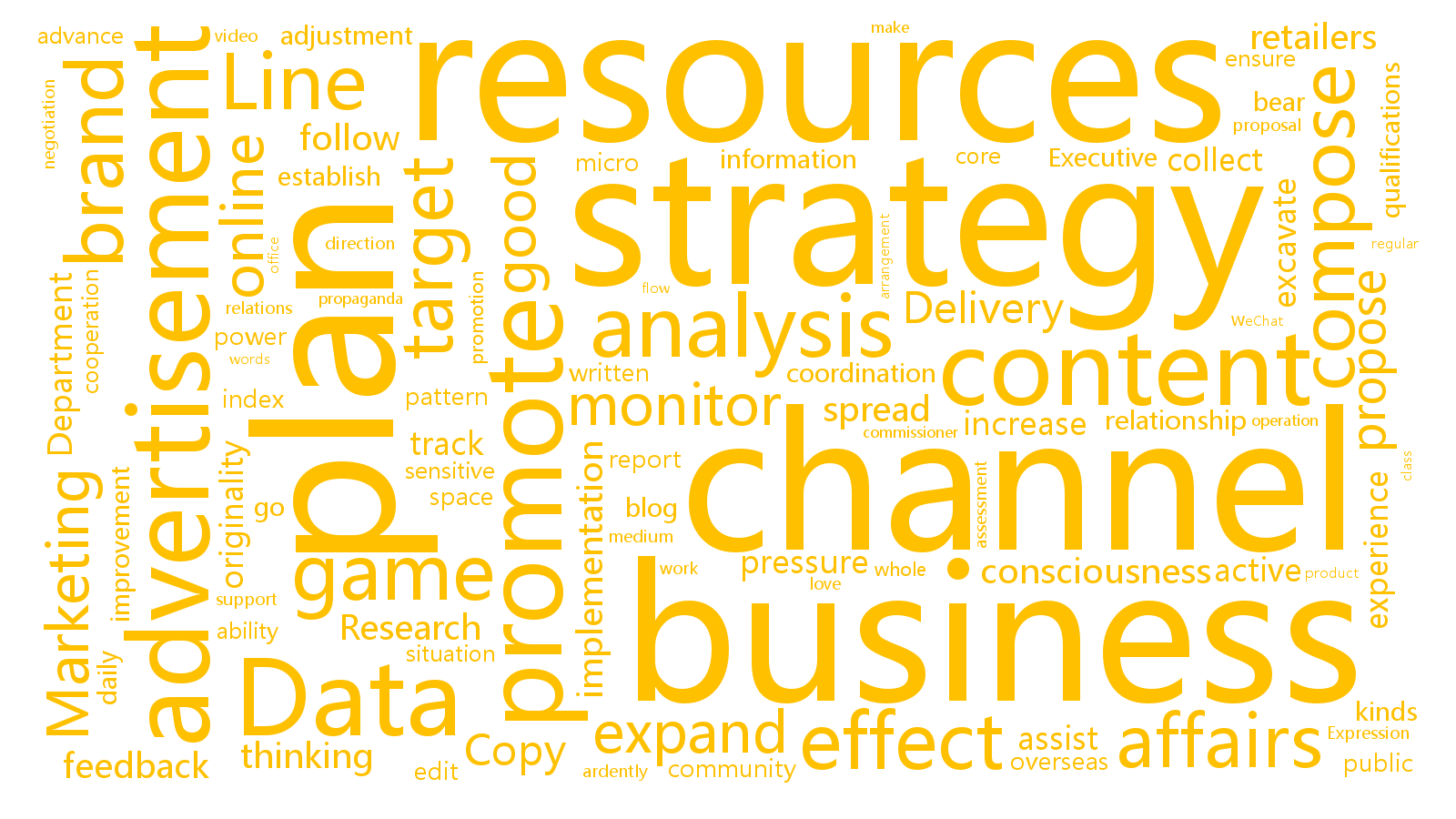}
\end{minipage}
}
\subfigure[state \#5] {
\begin{minipage}[b]{0.23\textwidth}
\includegraphics[width=1\textwidth]{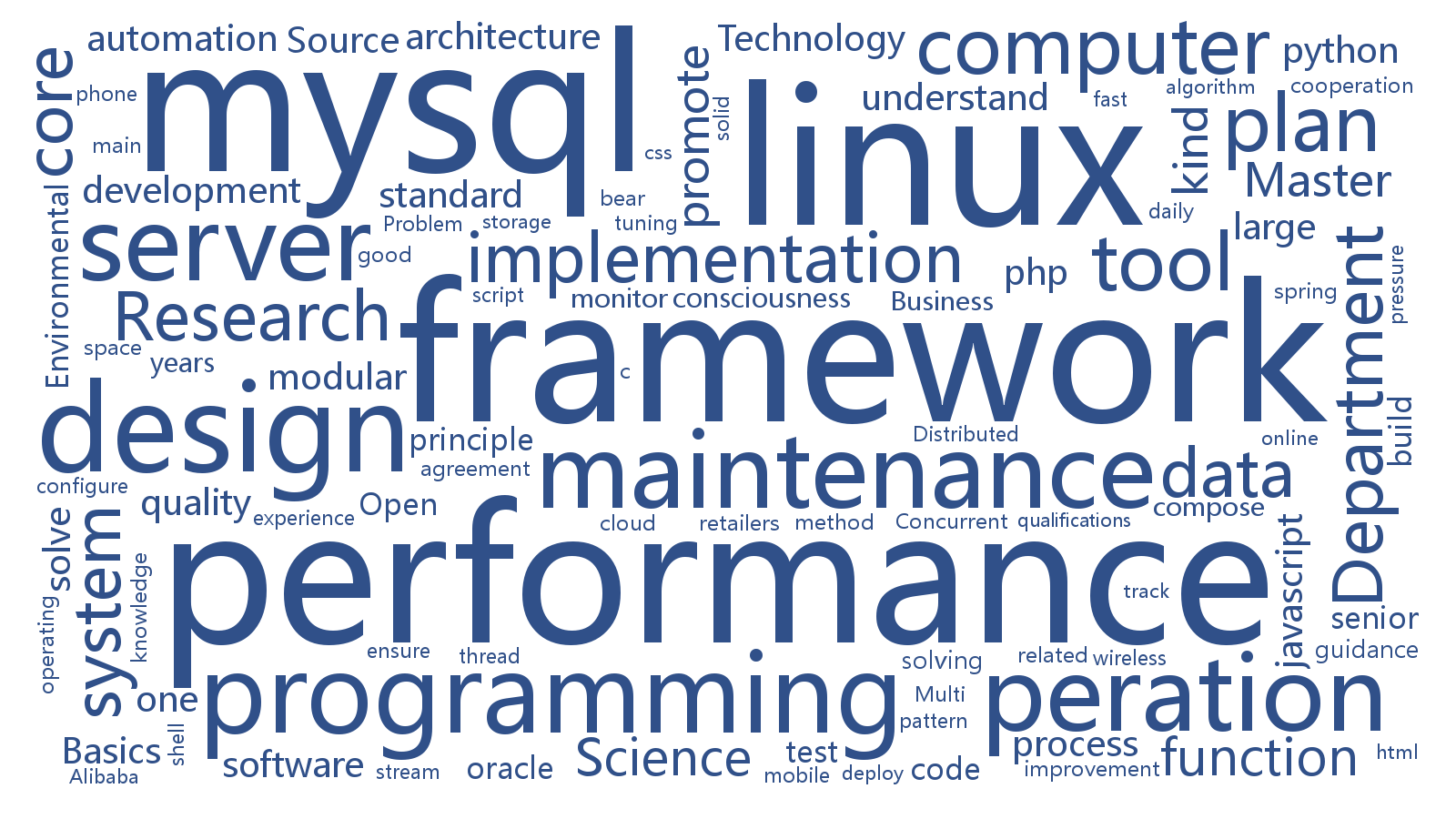} \\
\includegraphics[width=1\textwidth]{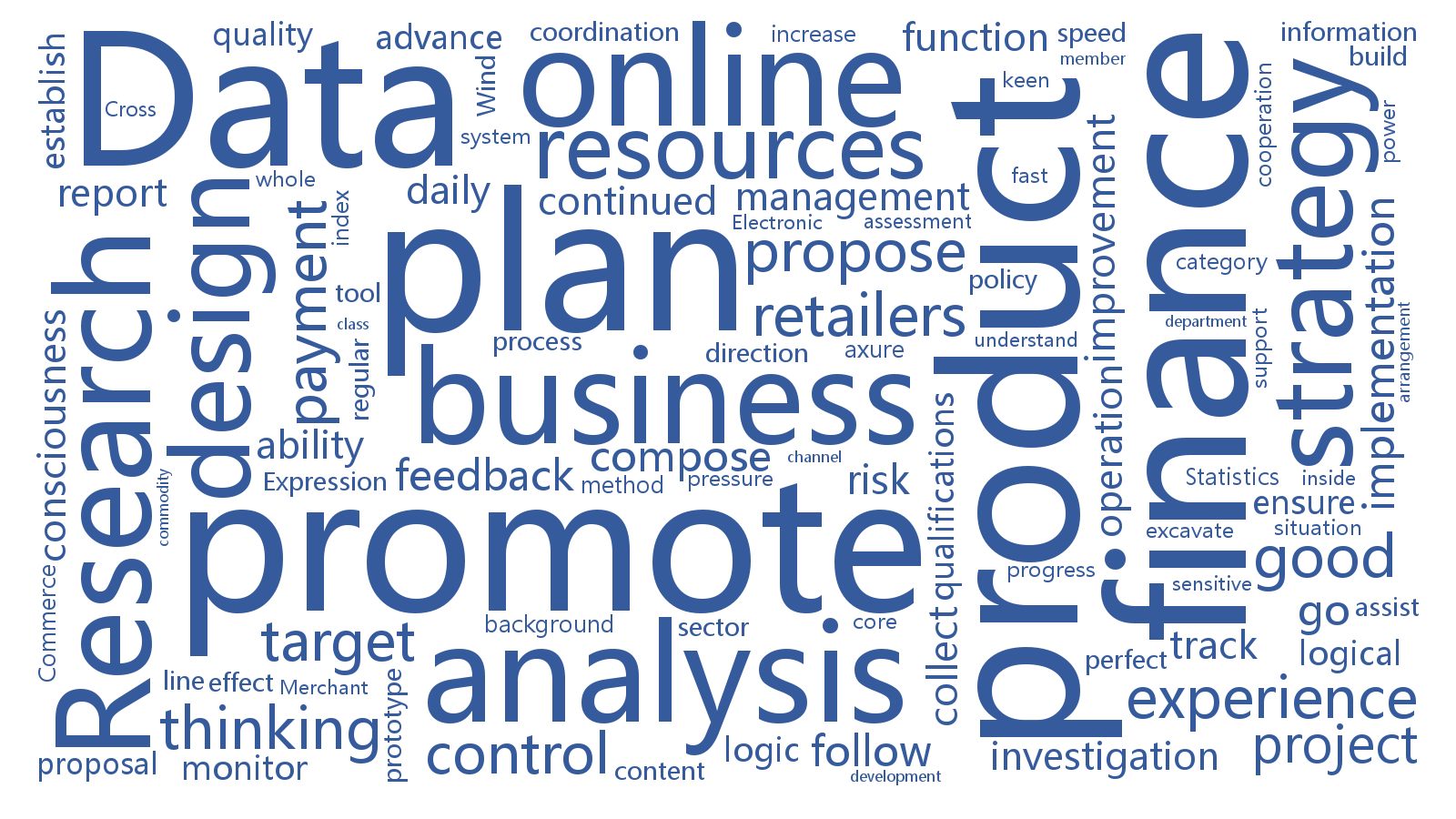} \\
\includegraphics[width=1\textwidth]{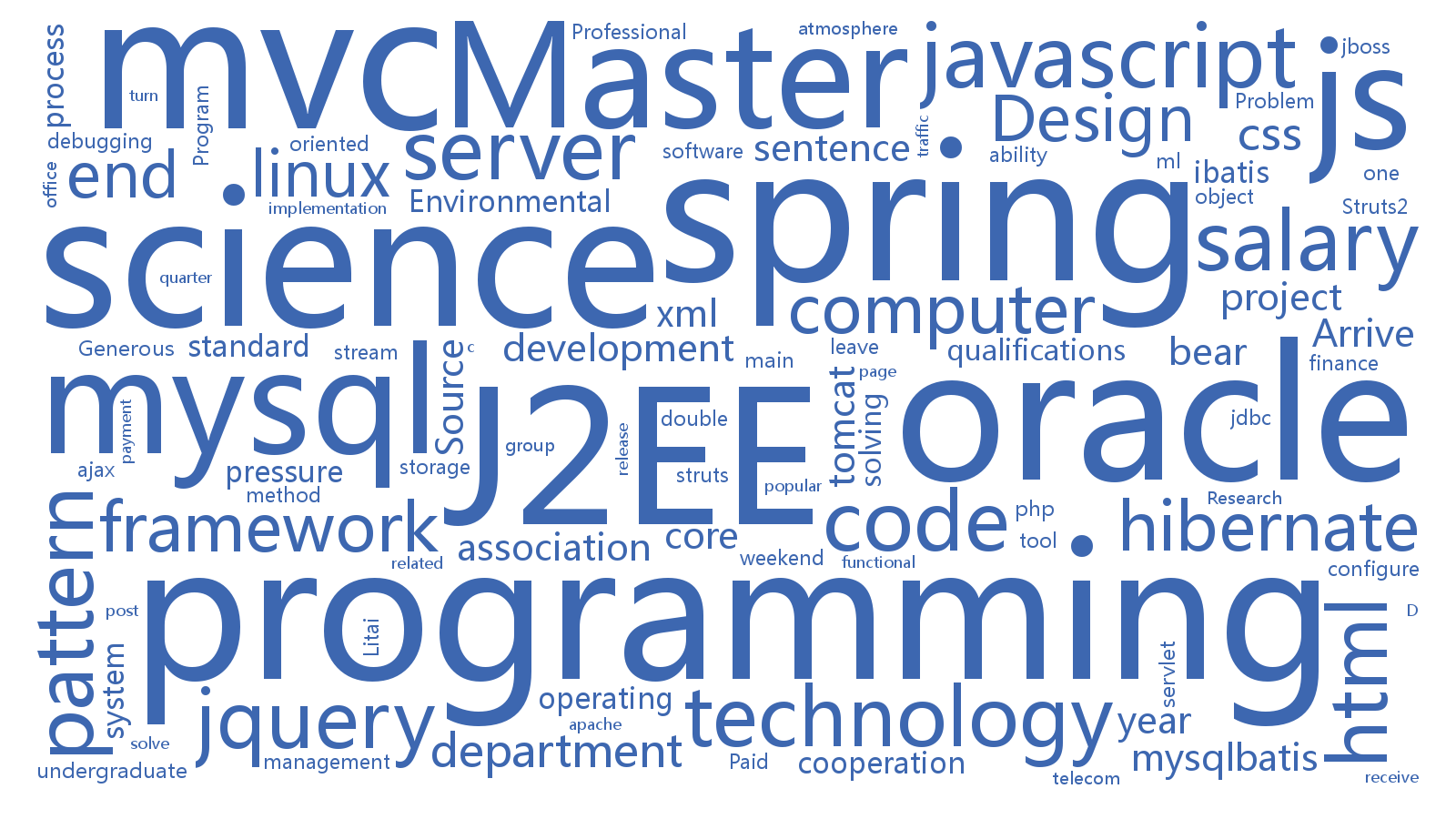} \\
\includegraphics[width=1\textwidth]{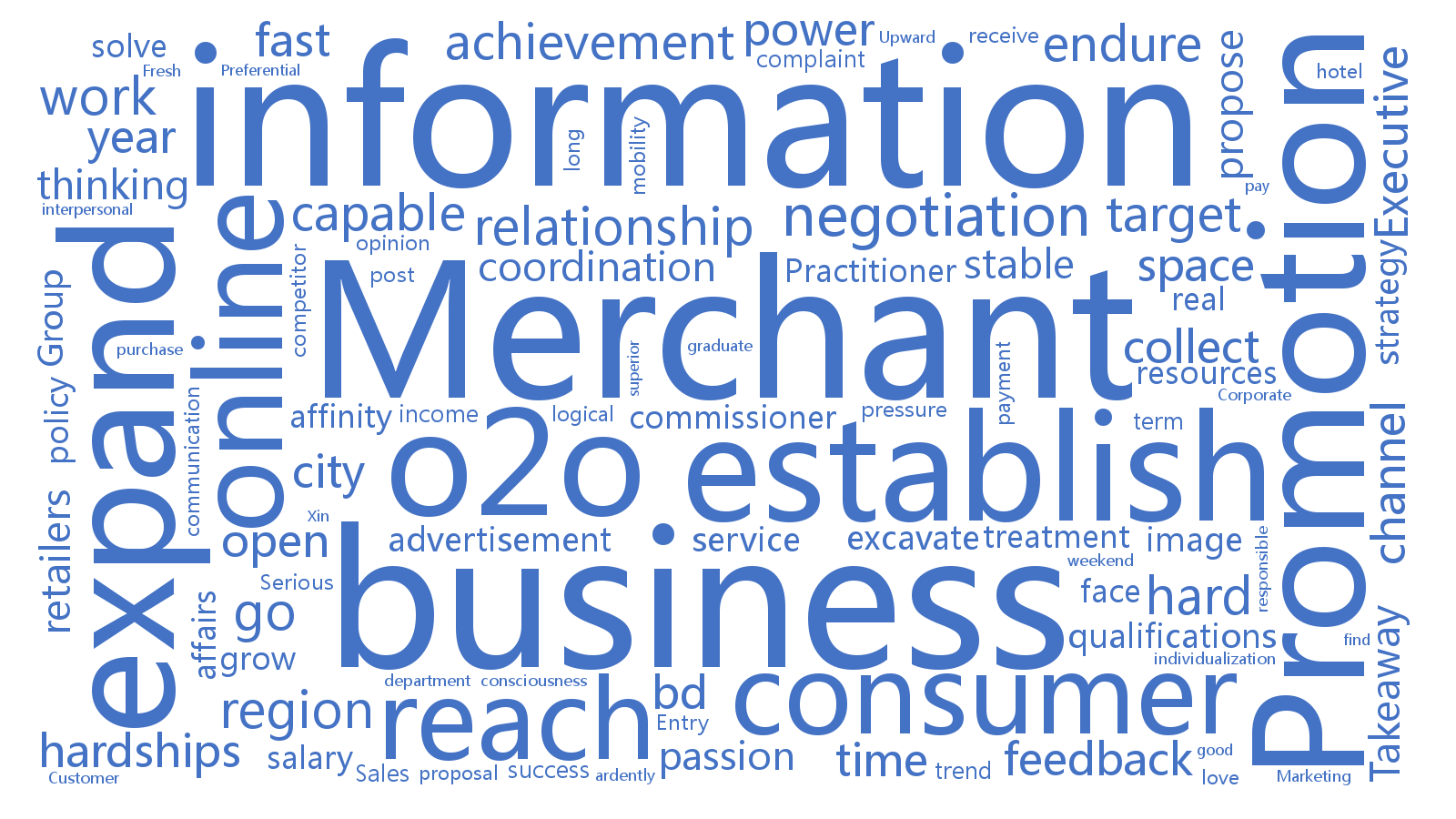}
\end{minipage}
}
\subfigure[state \#9] {
\begin{minipage}[b]{0.23\textwidth}
\includegraphics[width=1\textwidth]{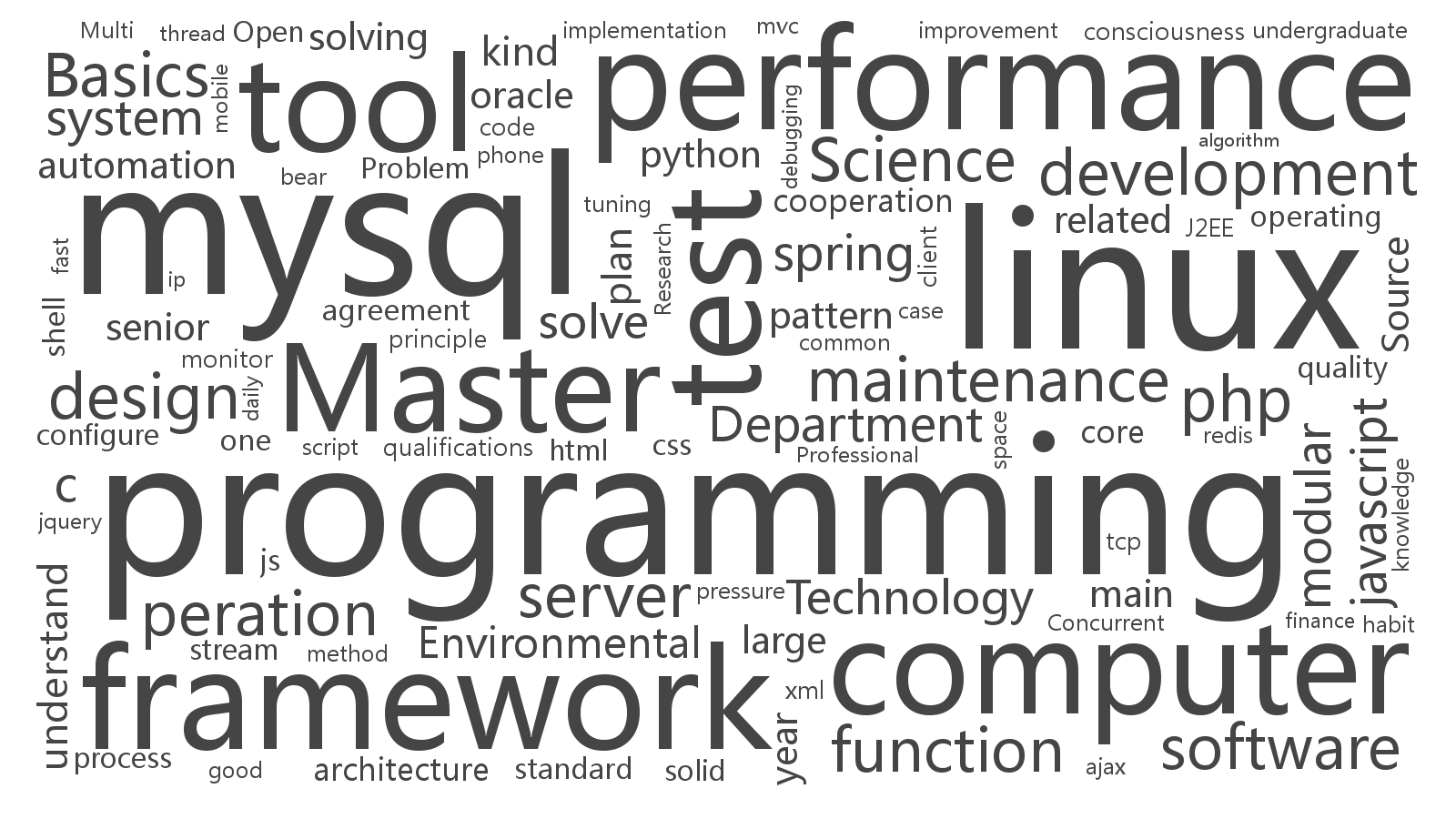} \\
\includegraphics[width=1\textwidth]{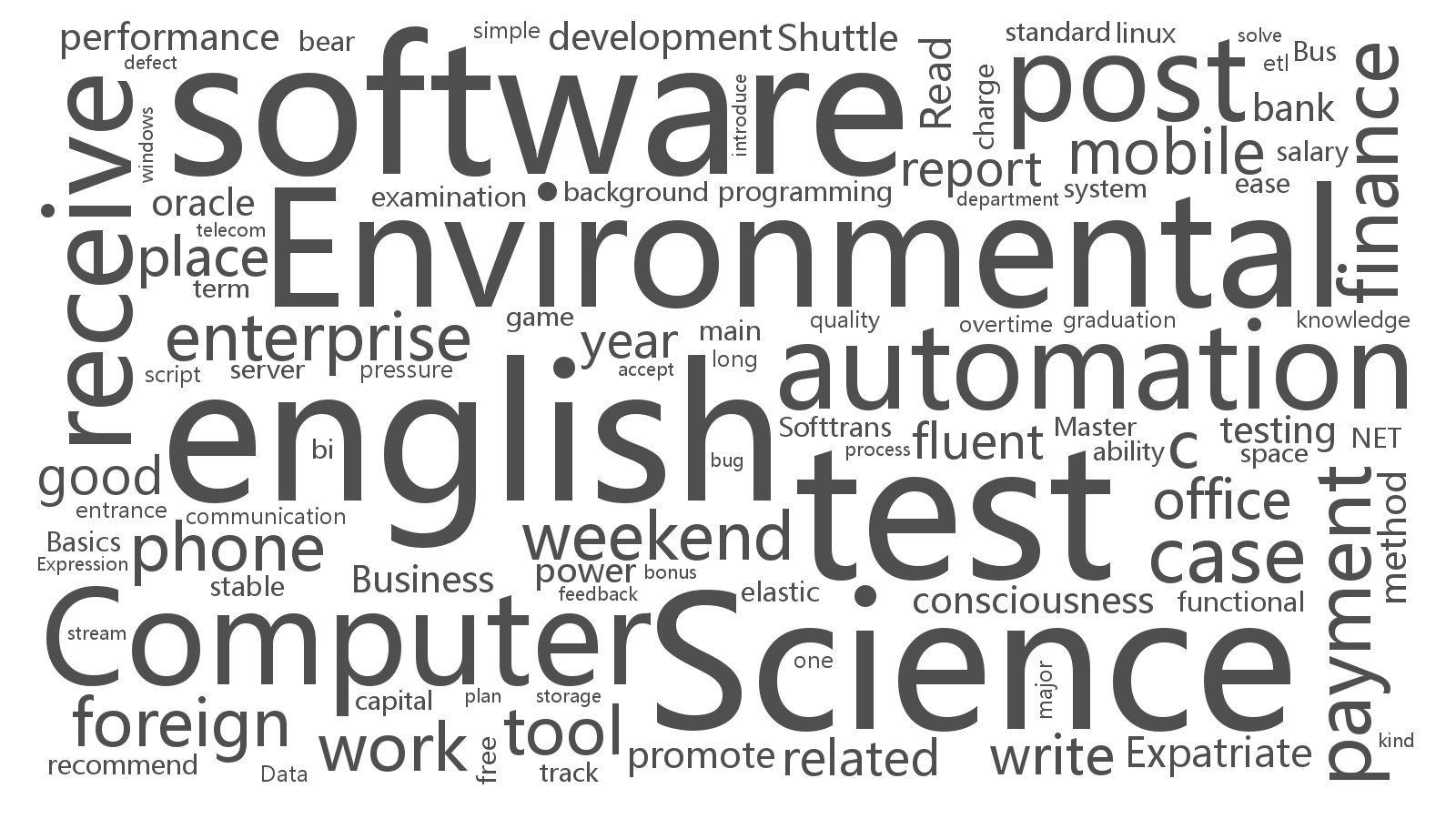} \\
\includegraphics[width=1\textwidth]{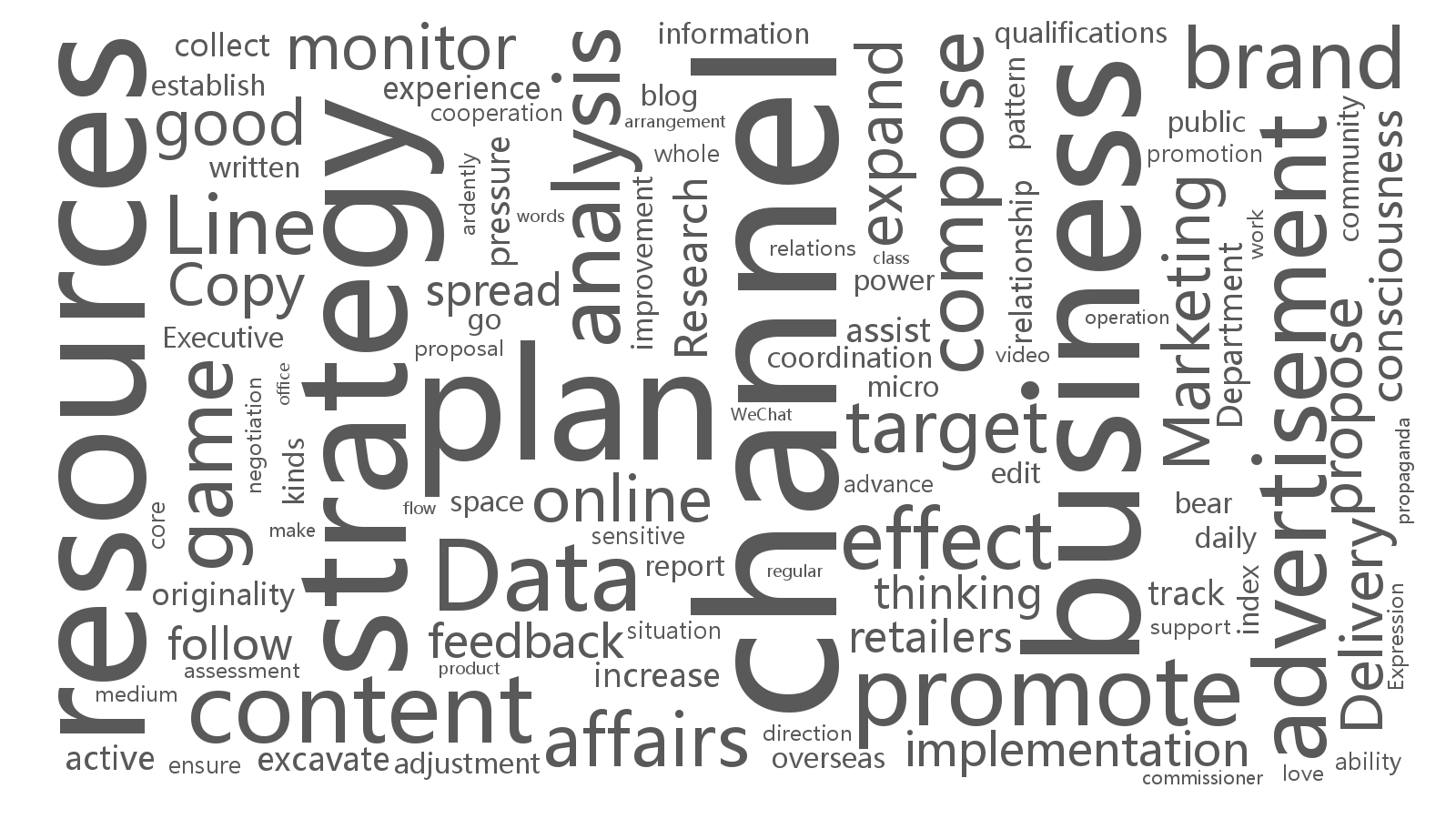} \\
\includegraphics[width=1\textwidth]{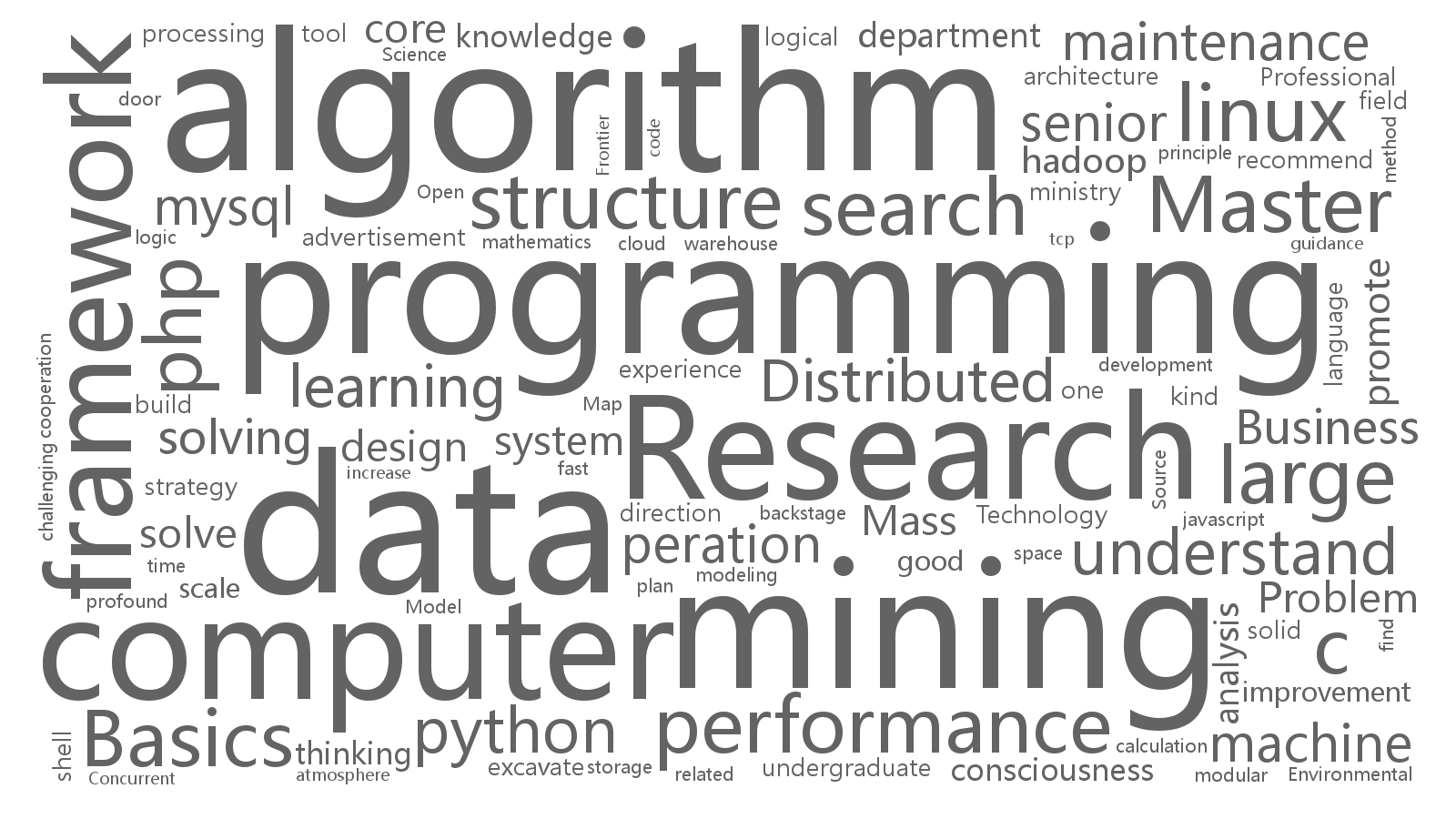}
\end{minipage}
}
\vspace{-2mm}
\caption{The word cloud representations of top 4 topics of selected recruitment states~(e.g., \#3, \#4, \#5, \#9), where the size of each keyword is proportional to its generative provability. The generative probabilities of these recruitment topics are shown in table~\ref{tab:prob_top_state}.}
\label{state&topic}
\vspace{-5mm}
\end{figure*}

Here, we will test whether the recruitment states discovered by our model are comprehensible. To solve this problem, we select 4 representative recruitment states according to above analysis and show their top 4 topics in figure~\ref{state&topic} by word cloud representations, where the larger words have higher generative probabilities. Meanwhile, Table~\ref{tab:prob_top_state} shows the corresponding generative probabilities of these topics.

We can find that the topic about ``programming'' always has very high probability in every state. In particular, the top \#1 topics in state \#1, state \#3, and state \#9 are the same. That means the demands for R\&D related positions are always exuberant, since the high-probability words, ``linux'' and ``mysql'', directly indicate the fundamental skill requirements of R\&D. Actually, the salary of software engineer has kept rising for a long time, which can support this discovery of our model. These states also show that the work about games and front-end are also very popular, which is consistent with our real-world observations.
Next, we further inspect states illustrated in figure~\ref{state&topic}.

\begin{table}[!t]
\vspace{-2mm}
 \caption{The probabilities of top recruitment topics of selected recruitment states~(e.g., \#1, \#3, \#4, \#9). The corresponding word cloud representations of these topics are shown in figure~\ref{state&topic}.}
 \centering
 \begin{tabular}{l r r r r}
  \toprule
   & state \#3 & state \#4 & state \#5 & state \#9 \\
  \midrule
    top \#1 & 0.22493     & 0.22172     & 0.18185   & 0.19441 \\
    top \#2 & 0.12399     & 0.17637     & 0.10126   & 0.12350 \\
    top \#3 & 0.10064     & 0.11360     & 0.07735   & 0.10018 \\
    top \#4 & 0.08932     & 0.08725     & 0.07242   & 0.08021 \\
  \bottomrule
 \end{tabular}\label{tab:prob_top_state}
\vspace{-2mm}
\end{table}

\vspace{-5mm}

\begin{itemize}
  \item \textbf{State \#3.} It is obvious that the top \#1 topic is about R\&D. The top \#3 topic, containing ``data'', ``analysis'', ``research'', and ``algorithm'', indicates the demand of recruiting senior researchers or algorithmic engineers. The top \#4 topic may be about propaganda (e.g., public relationship), because it contains several paperwork and advertising related words, such as ``compose'', ``edit'', ``WeChat'' and ``blog''(e.g., social network based advertising). Besides, what makes this state different is the top \#2 topic, which contains ``human'', ``resource'', ``office'', and ``assist''. All of them are about administrative management obviously. Actually, in our model, this topic only exists in this state. Since research development, propagandism, and administrative management are essential to any companies, we can conclude that this state covers the fundamental talent demands for most by high-tech companies. And figure~\ref{fig:state&time} also indicates state \#3 is the most popular.
  \item \textbf{State \#4.} This state is closely related to state \#5. We can find that the top \#2, \#3, and \#4 topics are relative normal. And the top \#1 topic, which contains ``data'', ``programming'', and ``algorithm'', is apparently about R\&D, too. However, what makes this topic different is word ``lbs''. Location-based Service~(LBS), such as map navigation or map localization, is the cornerstone of O2O, which is a kind of business model that uses online and mobile to drive offline local sales and become more and more popular since 2014. In figure~\ref{fig:state&time}, we notice that the popularity of this topic increased in 2014 and declined in 2015. This may be because the industry doorsill of this field is related high, so only a few large companies have capability to get into this field. Actually, only the largest IT companies in China, such as Baidu, Tencent, and sogou, provide such service now.
  \item \textbf{State \#5.} This is a very Chinese-style recruitment state. In top \#4 topic, we find word ``O2O'' as well as ``merchant'' and ``business''. Actually, with the proposal of ``Internet Plus''~\cite{int_plus}, thousands of start-up companies, focusing on O2O, sprung up across the country. Meanwhile, the others topics in this state are related normal in terms of technology. The top \#1 and \#3 topics are just about some popular programming language, web-framework, and database. It may indicate that ``O2O'' is just a business concept rather than a technology concept.
  \item \textbf{State \#9.} As shown in figure~\ref{fig:state&time}, this state exploded since February 2015. In the word cloud representations of its topics, the top \#4 is related meaningful. These high frequency words (``Baidu'', ``data'', ``mining'', ``large'', ``distributed'') indicate this is a big data related topic (e.g., large-scale machine learning and data mining). Its trend directly reveals that high-tech companies, such as Baidu, have paid more attention to big data related fields.
\end{itemize}

\subsubsection{Visualization of Trend over Companies}
\begin{figure*}[t]
\vspace{-2mm}
\begin{center}
\includegraphics[width=17cm]{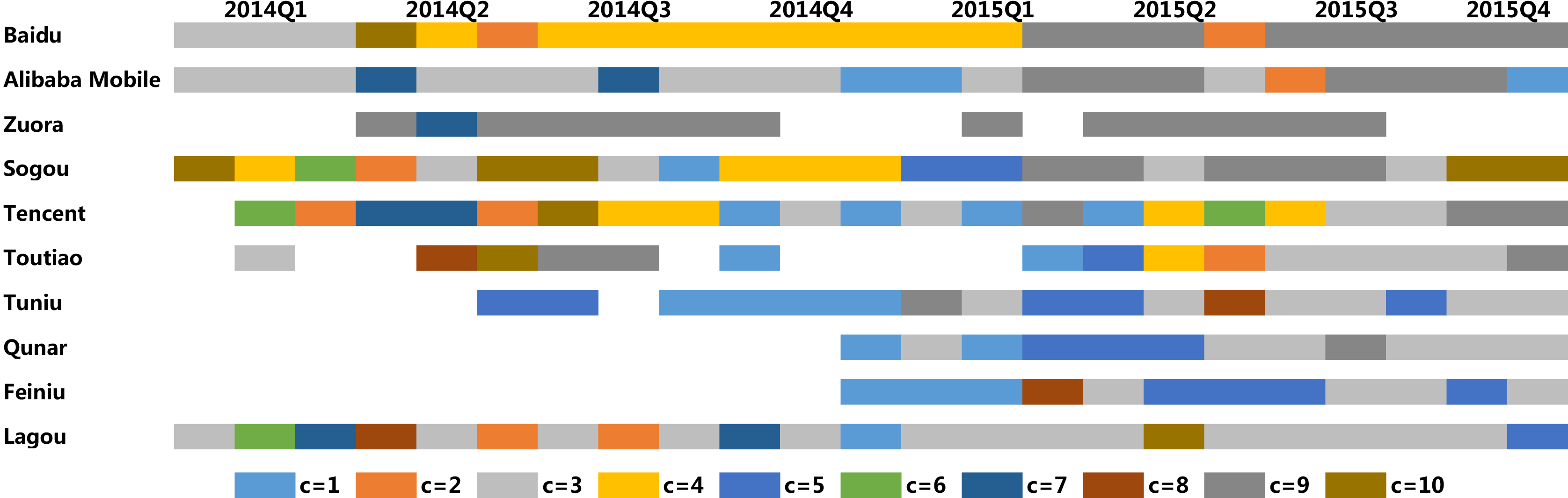}
\vspace{-2mm}
\caption{Visualization of the change of recruitment states over several companies, with different colors representing different states.}\label{fig:state_transfer}
\vspace{-8mm}
\end{center}
\end{figure*}

Here, we evaluate our model by checking the trend of several representative companies from a few important fields, such as Baidu is a famous high-tech company and Alibaba is the largest E-Commerce company in China. We visualize the evolution of the recruitment states of these companies in figure~\ref{fig:state_transfer}.

From figure~\ref{fig:state_transfer}, we can first observe that state \#3 is a very common state among most of companies. It is consistent with the analysis from topics of this state.
Besides, we find that state \#4, which is relevant to LBS, appears relatively frequently among Baidu, Sogou, and Tencent in 2014. Actually, all of these companies provide map service in China, especially Baidu. Baidu Map is the most popular navigation tool in China and many O2O companies use LBS API provided by Baidu to improve their services. So Baidu has remarkable strengths in LBS and has paid much attention to it indeed. Furthermore, Tencent, as one of the largest IT companies in China, its business is very scattered and covers many fields, such as game, social network, media, and entertainment. This kind of business strategy is directly reflected in Figure~\ref{fig:state_transfer}, where the recruitment state of Tencent changes frequently. Meanwhile, Baidu, Alibaba and Sogou~(another search engine) prefer state \#9, which is relevant to big data (e.g., large-scale machine learning and data mining), in 2015. Considering that their core business (search engine and E-commerce) has several data-related practical applications (advertising and recommender system), the preference is also reasonable.
In addition, we happened to find an interesting company, Zuora, whose state is almost state \#9. Actually, Zuora is an enterprise software company that aim to automate billing, commerce, and finance operations with a subscription business model. Such business model is naturally related to big data processing and thus has need for senior data-related talents.

Furthermore, we can observe that  state \#5, which is related to O2O, appears in company Tuniu, Qunar, and Feiniu frequently. Indeed, all of these companies aim to connect offline merchants and online customers, which is high consistent with O2O. Both of Tuniu and Qunar aim to provide one-stop travel booking service, such as hotel, ticket, car rental, and so on. And Feiniu is a B2C E-commerce wesite which is invested by a large retail corporation. The goal of its establishment is to fuse traditional offline service channel with online sale channel.

%****************************************************************************************************
\subsection{Evaluation of of Model Application}
%****************************************************************************************************
Here, we will evaluate the proposed model by predicting basic observations in future.

As introduced in Section~\ref{sec:model}, we make prediction by calculating the probability of basic observations with equation~\ref{equ:pred}. So in this experiment, we will compare the likelihood of overall observations in test set to prove the performance of our model. The test set is built up by the following approach. We extract the recruitment postings in November 2015, the companies of which had also released some jobs in October 2015. It means all of companies in test set have recruitment states in the previous time span. In the end, the test set contains 350 companies and 4002 recruitment postings. Besides, all of recruitment postings between January 2014 and October 2015 are treated as train data.

We select two model as baselines. One is dynamic Topic Model~(DTM)~\cite{blei2006dynamic}. DTM is a classic topic model for analyzing evolution of topics. It assumes topics evolve smoothly with respect to time and thus chains topics in adjacent epochs by state space models. Its performance was proved by predicting the next year of Science given all the articles from the previous years in~\cite{blei2006dynamic}. It is obvious that DTM cannot be used for prediction directly, but due to its assumption, the topics extracted by it can model future data better than static topic models in some cases. In this experiment, we follow this method to prove the assumption of DTM is not very solid in our problem and indicate the necessity of our model. The code of DTM was got from~\cite{code_DTM}. The number of topics is set to 100 and all of other parameters are set to default values.

In addition, we developed a simple sequence approach, called Byesian multivariate Hidden Markov Model~(B-mHMM), as the other baseline. Compared with MTLVM, this baseline associates states with words directly. The joint probability distribution of B-mHMM is as follows
\begin{equation}
\small
P(\chi,c|\rho,\iota)=\prod_{e,t}\Big(P(c_{e,t}|c_{e,t-1},\rho)\prod_{m,n}P(\chi_{e,t,m,n}|c_{e,t},\iota)\Big),
\end{equation}
where the first term is the same to equation~\ref{equ:chi&c}. And as shown in second term, we assume that given the recruitment state, all of observations $\chi_{e,t,m,n}$ are conditionally independent in B-mHMM and $P(\chi_{e,t,m,n}|c_{e,t},\iota)=Multi(\iota_{c_{e,t}})$. By this baseline, we aim to prove the latent hierarchical structure of our model are meaningful for modeling recruitment markets. Because of the similarity of B-mHMM and MTLVM, The details and inference of B-mHMM follow those of MTLVM and are omitted here.

Table~\ref{tab:prediction} shows the log likelihood of prediction with respect to different models. The larger number means better performance. Both of B-mHMM and MTLVM outperform DTM largely. It indicates it is reasonable to employ latent recruitment states to model the trend of recruitment markets. And the performance of MTLVM is also better than that of B-mHMM. All of these clearly validate the effectiveness of the proposed model.

\begin{table}[!t]
 \caption{The prediction performance of MTLVM and baseline methods in terms of log likelihood.}
 \centering
 \begin{tabular}{p{2.5cm} p{3.5cm}}
  \toprule
   & log likelihood \\
  \midrule
  MTLVM & \textbf{-2095737.793} \\
  B-mHMM & -2133504.721 \\
  DTM & -2599529.728 \\
  \bottomrule
 \end{tabular}\label{tab:prediction}
\end{table} 
\section{Related Work}
\label{sec:related}
%****************************************************************************************************
Generally, the related works of this paper can be grouped into two categories, namely recruitment market analysis, and sequential latent variable model.

\subsection{Recruitment Market Analysis}
Traditionally, recruitment market analysis can be regarded as a classic topic in human capital economics, which attracts generations of researchers to contribute ever since Adam Smith. From the Macro perspective, labor is always a crucial element in studying gross economy growth, money market, exchange market and the equilibrium~\cite{romer1996advanced}, ever since Solow proposed his growth model. Therefore, economists usually study topics about, for example, the demographic structure and participation rate of labor, the relation between inflation and unemployment rate, and how labor contributes in gross productivity or expenditures, etc~\cite{shapiro1984equilibrium}. In another Micro perspective, which is more relevant to our paper, all studies are set off from a basic market cleaning framework~\cite{varian2010intermediate}, where all employees choose their best balance between leisure and work, while all employers hire with budget constrain, and consequently the wage is derived as the marginal labor cost. Later researches improve our understandings by releasing constraints~\cite{hayashi2009nash} (e.g., acknowledging the market with segmentations/barriers and as non-cleaning), or by detailed investigations (e.g., forming better utility functions of employees, and studying the actions of employees with game theory).
Recently, several researchers in computer science try to employ data mining technology to solve these problems, such as offer categorization~\cite{malherbe2015bringing}, job skill analysis~\cite{litecky2012skills}.

However, previous research in economics efforts relies largely on the knowledge of domain experts or classic statistical models, and thus are usually too general to capture the high variability of recruitment market, and neglect the fine-grained market trend. On the other hand, the recent research in computer science still focuses on those traditional human resource problems. Therefore, in this paper we propose a new research paradigm for recruitment market analysis by leveraging unsupervised learning approach.

\vspace{2mm}
\subsection{Sequential Latent Variable Model}
\vspace{2mm}
Indeed, our novel sequential latent variable model MTLVM can be regarded as a combination of both Hidden Markov Model (HMM) and Hierarchical Dirichlet Processes~(HDP) within a Bayesian generative framework, which can intrinsically capture the sequential dependency and variability of latent variable (e.g., recruitment states and topics).

Specially, HMM based sequential latent variable models have been successfully applied to problems in a variety of fields, such as signal processing and speech recognition~\cite{rabiner1989tutorial,juang1991hidden}, biometrics~\cite{fredkin1992bayesian}, genetics~\cite{churchill1989stochastic,liu1999markovian}, economics~\cite{hamilton1989new}, and mobile Internet mining~\cite{huai2014toward,bao2012unsupervised,zhu2015popularity}. For much of their history, HMMs have been implemented by using recursive algorithms developed for parameter estimation~\cite{baum1970maximization}, which are viewed as ``black boxes'' by many statisticians. In recent years, some researchers proposed to use Bayesian methods to simulate HMM parameters from the posterior distribution, which can provide more scalable and stable process of parameter estimation for HMM~\cite{goldwater2007fully,guha2008bayesian}. Compared with the traditional maximum-likelihood estimation (MLE) based HMM learning solution, the Bayesian methods can directly maximize the probability of the hidden variables given the observed data by integrating over all possible parameter values rather than searching for an optimal set of parameter values. To this end, the model proposed in this paper also follows a Bayesian generative framework.

Latent Dirichlet Allocation (LDA)~\cite{blei2003latent} based latent variable models, have become one of the most powerful tools for mining textual data. However, most topic models~\cite{blei2006dynamic,zhang2015dynamic,zhu2015tracking} need a predefined parameter to indicate the number of topics, and thus fail to capture the variability of topics. To this end, the Hierarchical Dirichlet Processes~(HDP)~\cite{teh2006hierarchical} is proposed as an infinity version of topic model, which can automatically learn the number of topics. Therefore, in this paper we propose to ingrate HDP into our MTLVM for capturing the variability of latent recruitment topics.
%****************************************************************************************************

\section{Conclusion}
\label{sec:con}
%****************************************************************************************************
In this paper, we provided a large-scale data driven analysis of recruitment market trends. Specifically, we developed a novel sequential latent variable model, named MTLVM, which is designed for capturing the temporal dependencies of corporate recruitment states and is able to automatically learn the latent recruitment topics within a Bayesian generative framework. Moreover, to capture the variability of recruitment topics over time,  we designed hierarchical dirichlet processes for MTLVM. These processes allow to dynamically generate recruitment topics. Finally, we implemented a prototype system to empirically evaluate our approach based on large-scale real-world recruitment data. The results showed that our approach could effectively discover recruitment market trends and provide  guidances for both job recruiters and job seekers.
%****************************************************************************************************

\section{Appendix}
%****************************************************************************************************
Here, we will describe an analog of Chinese Restaurant Process~(CRP) for $P(\chi_{e,t}|\Lambda,c_{e,t})$ and corresponding inference in detail. Specifically, we first define $\{\psi_{e,t,m,z}\}_{z}$ as variables sampled from $G_{c_{e,t}}$ for each job posting and $|\psi_{e,t,m}|=Z_{e,t,m}$. Each $\phi_{e,t,m,n}$ is linked with a $\psi_{e,t,m,z}$ and $z_{e,t,m,n}$ is the index of $\psi_{e,t,m,z}$ for $\phi_{e,t,m,n}$. In other words, we have $\phi_{e,t,m,n}=\psi_{e,t,m,z_{e,t,m,n}}$. Besides, let $i_{e,t,m,z}$ denote the number of $\{\phi_{e,t,m,n}\}_n$ linked with $\psi_{e,t,m,z}$. According to CRP, we can integrate out the $D_{e,t,m}$ and get the conditional distribution of $\phi_{e,t,m,n}$ as follows.
\begin{eqnarray}
\nonumber \phi_{e,t,m,n}|\phi_{e,t,m,1},...,\phi_{e,t,m,n-1},\{G_c\}_{c},c_{e,t},\gamma_2 \sim \\
\sum_{z=1}^{Z_{e,t,m}}\frac{i_{e,t,m,z}}{n-1+\gamma_2}\delta_{\psi_{e,t,m,z}}+\frac{\gamma_2}{n-1+\gamma_2}G_{c_{e,t}},
\label{equ:phi}
\end{eqnarray}
where $\delta_{\psi}$ is a probability measure concentrated at $\psi$.

If a new $\psi$ is sampled, it indicates the second term on the right-hand side of equation~\ref{equ:phi} is chosen. Then we need to add $Z_{e,t,m}$ by 1, sample a new $\psi$ by equation~\ref{equ:psi}, and allocate $\phi_{e,t,m,n}$ to it. If the sampled value is an existed $\phi$, we just need allocate $\phi_{e,t,m,n}$ to it.

Second, we define $\{\eta_{c,s}\}_{s}$ as variables sampled from $G_0$ for each recruitment state and $|\eta_{c}|=S_c$. Each $\psi_{e,t,m,z}$ is linked with a $\eta_{c,s}$ and $s_{e,t,m,z}$ is the index of $\eta_{c,s}$ for $\psi_{e,t,m,z}$. In other words, $\psi_{e,t,m,z}=\eta_{c,s_{e,t,m,z}}$. Besides, let $j_{c,s}$ denote the number of $\{\psi_{e,t,m,z}\}_{e,t,m}$, which is linked with $\eta_{c,s}$ and $c_{e,t}=c$. Similarly, we can integrate out $G_c$ and get the conditional distribution of $\psi_{e,t,m,z}$ as follows.
\begin{eqnarray}
\nonumber & \psi_{e,t,m,z}|\psi_{1,1,1,1},...,\psi_{e,t,m,z-1},Q_0,\gamma_1 \sim \\
& \sum_{s=1}^{s_c}\frac{j_{c,s}}{\sum_{s=1}^{s_c}j_{c,s}+\gamma_1}\delta_{\eta_{c,s}}+\frac{j_{c,s}}{\sum_{s=1}^{s_c}j_{c,s}+\gamma_1}Q_0.
\label{equ:psi}
\end{eqnarray}
The sampling process is similar to $\phi$.

Third, we let $\{\theta_{k}\}_k$ denote the variables sampled from $H$ and $|\theta|$ is $K$. Each $\eta_{c,s}$ is linked with a $\theta_{k}$ and $k_{c,s}$ is the index of $\theta_{k}$ for $\eta_{c,s}$, i.e., $\eta_{c,s}=\theta_{k_{c,s}}$. And we also let $o_{k}$ denote the number of $\{\eta_{c,s}\}_{c,s}$. Now we can write the conditional distribution of $\eta_{c,s}$ directly as
\begin{eqnarray}
\nonumber & \eta_{c,s}|\eta_{1,1},...,\eta_{c,s-1},H,\gamma_0 \sim \\
& \sum_{k=1}^{K}\frac{o_k}{\sum_{k=1}^{K}o_k+\gamma_0}\delta_{\theta_{k}}+\frac{o_k}{\sum_{k=1}^{K}o_k+\gamma_0}H.
\label{equ:eta}
\end{eqnarray}

Next, we will describe a Gibbs sampling method yielded from above. Specifically, we follow the inference method in~\cite{teh2006hierarchical} and sample $z$, $s$, $k$ and $\theta$, rather than dealing with $\phi$, $\psi$ and $\eta$ directly.

\textbf{Sampling $z$, $s$, and $k$. }Relying on equation~\ref{equ:phi}, we can easily compute the conditional distribution of $z_{e,t,m,n}$ by
\begin{eqnarray}
\nonumber P(z_{e,t,m,n}=z|z^{-e,t,m,n},s,k,\theta,\chi) \propto \\
\begin{cases}
\gamma_2 P(\chi_{e,t,m,n}|\theta_{k_{e,s_{e,t,m,z}}}) & \text{new}, \\
i_{e,t,m,z}^{-e,t,m,n}P(\chi_{e,t,m,n}|\theta_{k_{e,s_{e,t,m,z}}}) & \text{used},
\end{cases}
\label{equ:z}
\end{eqnarray}
where $i_{e,t,m,z}^{-e,t,m,n}$ is $i_{e,t,m,z}$ except the variable $\phi_{e,t,m,n}$. The likelihood of $z_{e,t,m,n}=z$ is simply $P(\chi_{e,t,m,n}|\theta_{k_{e,s_{e,t,m,z}}})$ given all of other variables. And the prior probability that $z_{e,t,m,n}$ samples an existed $\psi_{e,t,m,z}$ is proportional to $i_{e,t,m,z}^{-e,t,m,n}$ is $i_{e,t,m,z}$.  Its prior probability for a new $\psi$ is proportional to $\gamma_{2}$. The process of sampling $s$ and $k$ is similar to that of sampling $z$.

\textbf{Sampling $\theta$.} Given $z$, $s$, and $k$, $\{\theta_{k}\}_{k}$ are mutually independent. So the conditional distribution for each $\theta_k$ is only related with all of $\chi$ that linked with it. It follows
\begin{eqnarray}
\nonumber P(\theta_{k}|z,t,k,\theta^{-k},\chi) \propto h(\theta_{k})\cdot ~~~~~~~~~~~~~~~~~~~~~~~~~~\\
\resizebox{.9\hsize}{!}{$\prod_{c,s:k_{c,s}=k}\prod_{e,t:c_{e,t}=c}\prod_{m,z:s_{e,t,m,z}=s}\prod_{n:z_{e,t,m,n}=z}P(\chi_{e,t,m,n}|\theta_{k})$},
\end{eqnarray}
where $h(\theta_{k})$ is the density of measure $H$ at parameter $\theta_{k}$.

% \small
\bibliographystyle{abbrv}
\bibliography{cite}
\end{document}